\pdfoutput=1

\documentclass[11pt]{article}

\usepackage{acl}

\usepackage{times}
\usepackage{latexsym}
\usepackage{booktabs}
\usepackage{multirow}
\usepackage{tabularx}
\usepackage{graphicx}
\usepackage{listings}
\usepackage{longtable}
\usepackage{xtab}
\usepackage{makecell}
\usepackage{cleveref}
\usepackage[T1]{fontenc}

\usepackage[utf8]{inputenc}

\usepackage{microtype}

\usepackage{inconsolata}

\title{Python is Not Always the Best Choice: Embracing Multilingual Program of Thoughts}

\author{Xianzhen Luo$^1$, Qingfu Zhu$^1$\thanks{\ \ Corresponding author}, Zhiming Zhang$^1$, Libo Qin$^2$, \\ 
\textbf{Xuanyu Zhang}$^3$, \textbf{Qing Yang}$^3$, \textbf{Dongliang Xu}$^3$, \textbf{Wanxiang Che}$^1$ \\
$^1$Harbin Institute of Technology, Harbin, China\\
$^2$Central South University, Changsha, China\\
$^3$Du Xiaoman (Beijing) Science Technology Co., Ltd.\\
\texttt{\{xzluo, qfzhu, zmzhang, car\}@ir.hit.edu.cn} \\
\texttt{lbqin@csu.edu.cn} \\
\texttt{\{zhangxuanyu, yangqing, xudongliang\}@duxiaoman.com}
}
\begin{document}
\maketitle
\begin{abstract}

Program of Thoughts (PoT) is an approach characterized by its executable intermediate steps, which ensure the accuracy of the logical calculations in the reasoning process.
Currently, PoT primarily uses Python. However, relying solely on a single language may result in suboptimal solutions and overlook the potential benefits of other programming languages. In this paper, we conduct comprehensive experiments on the programming languages used in PoT and find that no single language consistently delivers optimal performance across all tasks and models. The effectiveness of each language varies depending on the specific scenarios.
Inspired by this, we propose a task and model agnostic approach called MultiPoT, which harnesses strength and diversity from various languages. 
Experimental results reveal that it significantly outperforms Python Self-Consistency. Furthermore, it achieves comparable or superior performance compared to the best monolingual PoT in almost all tasks across all models. In particular, MultiPoT achieves more than 4.6\% improvement on average on ChatGPT (gpt-3.5-turbo-0701)\footnote{Code and data are released at \url{https://github.com/Luowaterbi/MultiPoT}}.
\end{abstract}

\section{Introduction}

\begin{figure}[!th]
\centering
\includegraphics[width=\columnwidth]{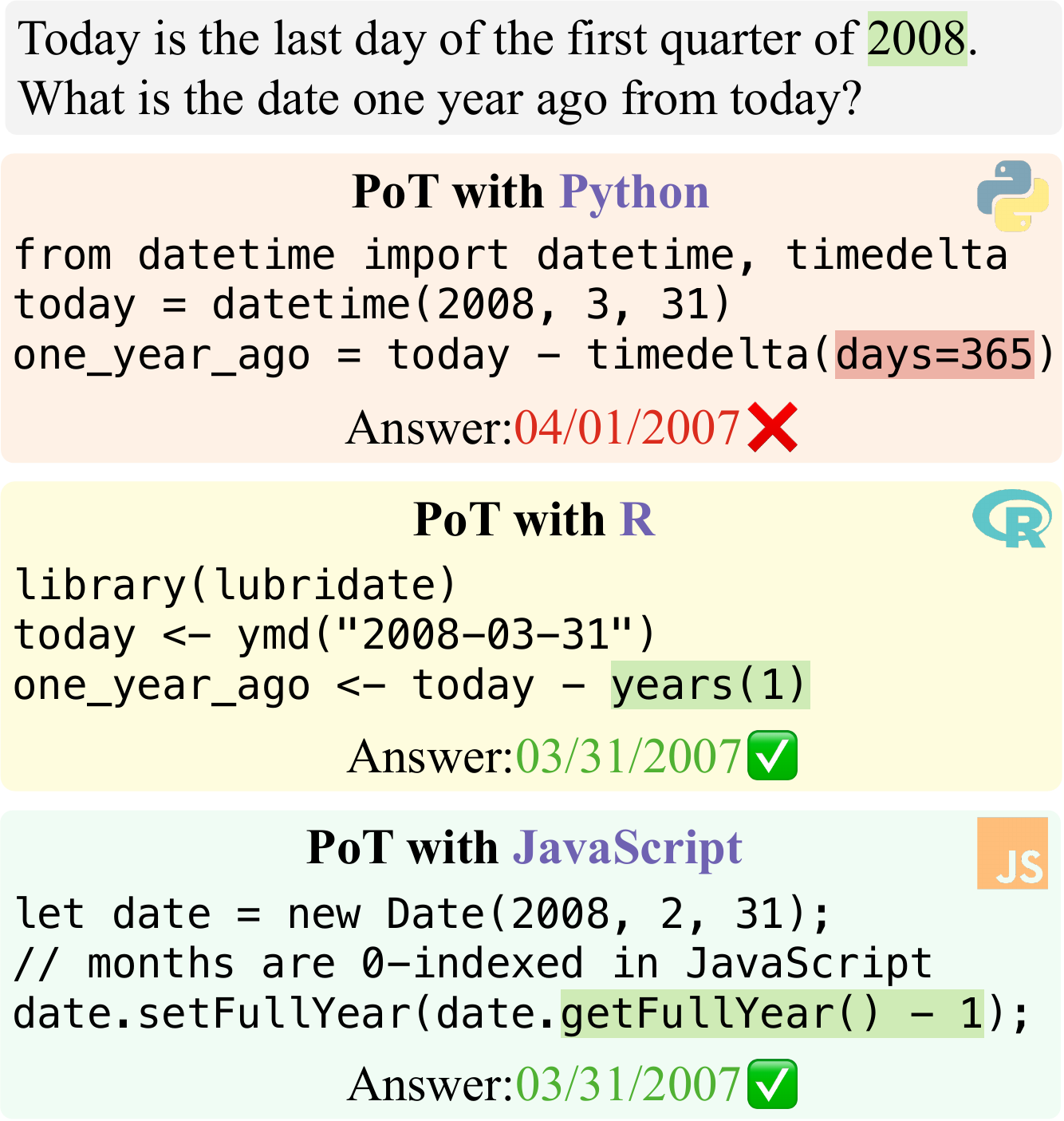}
\caption{Comparison of PoT with different PLs. 
Python's `\texttt{timedelta}' lacks support for year computation, leading to a leap year (2008 has 366 days) error by subtracting 365 days. R and JavaScript directly compute the year and get the correct answer.
}
\label{fig:comparison}
\end{figure} 
Program of Thoughts (PoT) aims to prompt Code Large Language Models (Code LLMs) to decompose complex problems into successive executable codes~\cite{Pal, PoT}. 
Through execution by an external interpreter, the final results are accurately obtained, decoupling the computational process from the LLMs.
PoT significantly reduces computation errors and improves reasoning performance~\citep{wang2023exploring}.
Subsequently, benefiting from its flexibility and scalability, it is gradually applied to a broader spectrum of fields like image reasoning~\cite{vipergpt,visualprogramming}, financial QA~\cite{bizbench} and robotic control~\cite{CoC}. Nowadays, PoT has become a key method for enabling intelligence in agents~\citep{IA-Survey, codeact}. The widespread applicability highlights its significance.

Despite significant progress, PoT has a notable limitation: to the best of our knowledge, \textbf{all research on PoT focuses on Python}. However, since Code LLMs are capable of multilingual generation,\footnote{In this paper, our ``multilingual'' represents multiple programming languages, not natural languages.} and most of the reasoning tasks are language-independent, many other programming languages (PLs) can also be applied to PoT, especially when considering their unique strength and diversity. 
From the perspective of \textbf{tasks}, different PLs represent PoT in different forms. As shown in Figure~\ref{fig:comparison}, the representation and calculation of dates in R is more concise than that in Python, which can reduce the complexity when LLMs generate PoTs. 
From the perspective of \textbf{models}, their multilingual ability is inconsistent. For instance, C++ of Deepseek Coder outperforms Python on the code generation task~\cite{deepseek-coder}. It is natural to wonder whether this phenomenon also occurs on reasoning tasks.
Therefore, a crucial question is raised with these perspectives: \emph{Is Python truly the optimal language for all tasks and models for PoT?} 
Relying on Python may lead to a local optimum. 
In Figure~\ref{fig:comparison}, 
Python's `\texttt{timedelta}' does not support `year', resulting in a miscalculation for the leap year. In contrast, R and JavaScript yield the correct answer.
Motivated by this, we conduct comprehensive experiments for multilingual PoTs.
Beyond Python, we select four PLs: three widely used general languages (JavaScript, Java, and C++) and a niche but comprehensive language (R).
For a comprehensive comparison, we identify five distinct sub-tasks within reasoning tasks: math applications~\cite{gsm8k,SVAMP,ASDIV}, math~\cite{MATH}, tabular, date, and spatial~\cite{BBH-hard}. 
We select four backbone LLMs: ChatGPT (gpt-3.5-turbo-0701) and three strongest Code LLMs (Starcoder~\cite{Starcoder}, Code Llama~\cite{code-llama}, and Deepseek Coder~\cite{deepseek-coder}).
Under both greedy decoding and Self-Consistency~\cite{self-consistency} settings,
we answer that ``\textit{\textbf{Python is not always the optimal choice, as the best language depends on the specific task and model being used.}}''

In addition to the analysis contribution, to \textbf{leverage the strength of multiple PLs}, we further introduce a simple yet effective approach, called \textbf{MultiPoT} (\textbf{Multi}lingual \textbf{P}rogram \textbf{o}f \textbf{T}houghts). 
It is a task and model agnostic approach, which uses LLMs to synchronously generate PoTs with various PLs and subsequently integrates their results via a voting mechanism. 
The use of \textbf{multiple PLs also provides greater diversity} and reduces the probability of repeating the same errors compared to single-language sampling.
Experimental results demonstrate that MultiPoT outperforms Python Self-Consistency significantly.
Furthermore, MultiPoT effectively matches or even surpasses the top-performing languages across nearly all tasks and models, and outperforms on averages.
Especially on both ChatGPT and Starcoder, MultiPoT performs the best on four out of five tasks, with only a slight underperformance on the remaining task, and shows an improvement of over 4.6\% compared to the best monolingual PoT on average.

Our contributions are summarized below:
\begin{itemize}
    \item We conduct comprehensive experiments of PoTs with different PLs across various reasoning tasks and models, revealing that the choice of PL is dependent on tasks and models.
    \item We introduce a task and model agnostic approach called MultiPoT, which integrates multilingual PoTs and leverages strength and diversity across various PLs.
    \item Experimental results show that MultiPoT outperforms Python Self-Consistency and matches or surpasses the best language of each scenario. On both the model and task averages, MultiPoT enhances performance.
\end{itemize}

\section{Related Work}
\subsection{Program of Thoughts}

CoT is a specific form of in-context learning~\cite{CoT, gpt3, palm}. Its demonstrations consist of intermediate steps imitating the human thought process. It significantly enhances model's reasoning capabilities~\cite{yang2023large} but suffers from errors associated with calculations~\cite{madaan2022text}. 
CoT always uses Self-Consistency~\cite{wang2023selfconsistency} to improve answer accuracy through sampling and voting.

PoT~\cite{PoT, Pal} is an extension of CoT to avoid incorrect calculation. It represents intermediate steps as comments and code and executes the entire program with an interpreter to obtain answers.
PoT not only excels in reasoning tasks but has rapidly extended to practical applications, including chart understanding, image reasoning, financial QA and robotic control~\cite{tinychart,vipergpt,visualprogramming,bizbench,CoC}. It has become a key method for agents to perform complex reasoning and tool invocation~\cite{IA-Survey, codeact}.
It is important to note that all previous PoT work only use Python. For the first time, we are exploring PoTs that use multiple PLs.

\begin{figure*}[t]
\centering
\includegraphics[width=\textwidth]{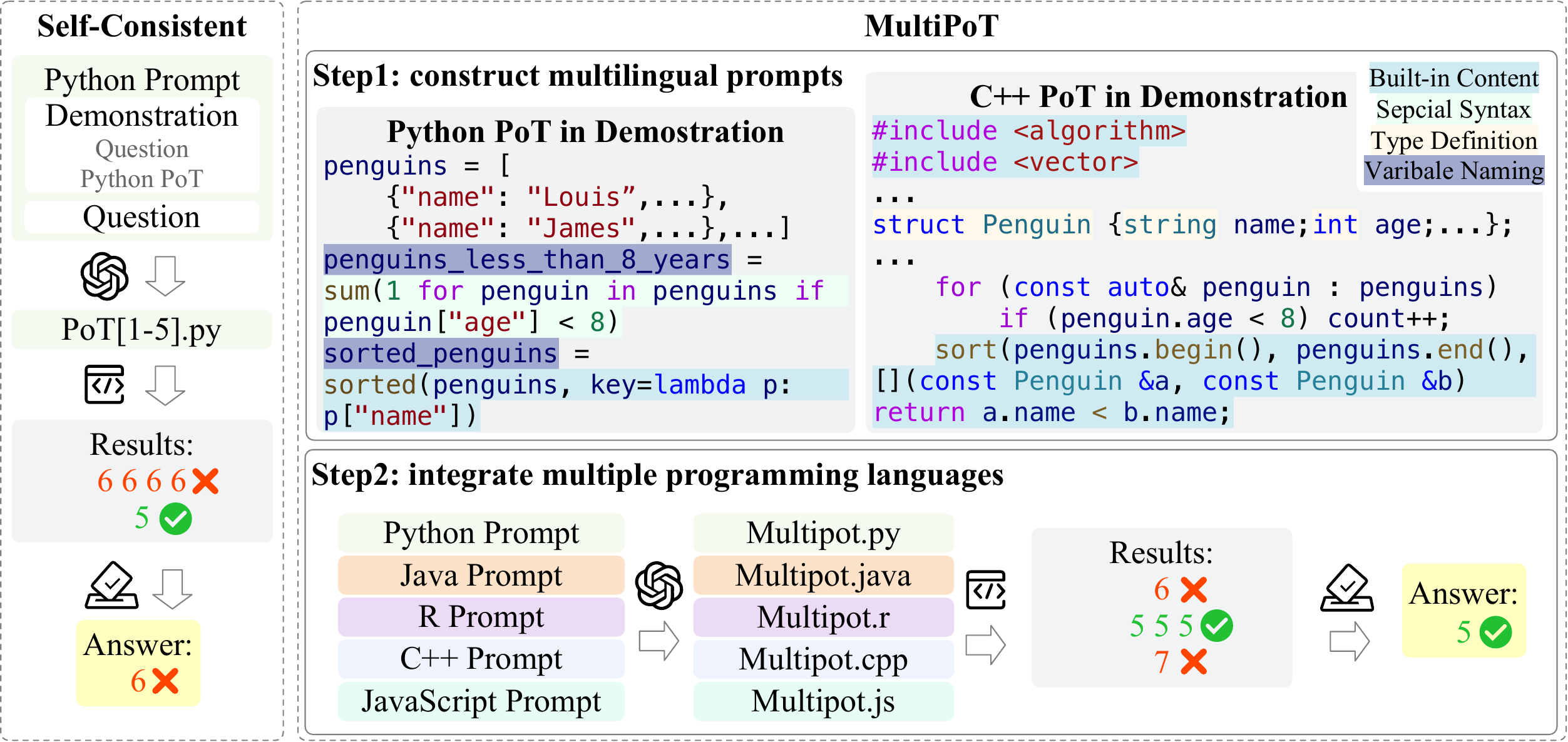}
\caption{
An overview of MultiPoT and Self-Consistency. MultiPoT first constructs prompts for each PL, ensuring a consistent reasoning process while also considering the distinct coding styles.
It then integrates these PLs: generating multilingual PoTs based on the prompts, executing them to gather results, and finally voting for the answer. In contrast to Self-Consistency’s single-language focus, MultiPoT leverages multiple PLs.
}
\label{fig:multipot}
\end{figure*}

\subsection{Usage of Multiple PLs}
The training datasets naturally include a variety of PLs, endowing Code LLMs with the ability to handle multilingual programming~\cite{Stack, vault, pile, codegen, codex}. This capability extends code tasks like generation, optimization, translation, and repair to other languages beyond Python~\cite{alphacode2, shypula2023learning, 10.1145/3611643.3616350, Wu_2023}.
Despite the progress, current multilingual research~\cite{jin2023inferfix, Repair, khare2023understanding} mainly focuses on code-related tasks, neglecting the potential of PLs as tools to assist in other tasks. Additionally, these studies often treat each language separately without interaction. 
Our study pioneers the use of multiple PLs in reasoning tasks and introduces a novel integrated approach, leveraging the collective strength and diversity of various PLs to enhance overall performance.


\section{Methodology}

Figure~\ref{fig:multipot} provides an overview of MultiPoT and Self-Consistency to highlight their differences. 
Concretely, MultiPoT consists of two main steps.
First, a dedicated prompt is designed for each PL to sufficiently leverage the capability of the model with regard to the PL (Section~\ref{sec:construct}).
Second, PoTs in various PL are respectively generated by prompting the LLM with the prompts. 
The final answer is obtained by executing the PoTs and integrating their results via a voting mechanism (Section~\ref{sec:intergrate}).
Distinct from Self-Consistency, which relies on a single PL, MultiPoT integrates various PLs to utilize their strength and diversity.

\subsection{Multilingual Prompts Construction}
\label{sec:construct}

To instruct a LLM to generate PoT for a given question, a demonstration is included in the prompt.
The demonstration consists of an example question and PoT.
To ensure fairness, demonstrations of various PLs share the same example questions.
Based on that, to efficiently leverage the capability of a LLM with regard to a PL, each PL is provided with a dedicated example PoT, taking into account its language-specific characteristics~\cite{wang-etal-2023-code4struct}.
Note that language-agnostic features, such as algorithms and data structures, remain the same for example PoTs of all PLs, ensuring an identical reasoning process.



Concretely, the language-specific characteristics of each PL for constructing its dedicated example PoT includes \textbf{Built-in Content}, \textbf{Special Syntax}, \textbf{Type Definition}, and \textbf{Varibale Naming}.
Figure~\ref{fig:multipot} provides some examples of the characteristics. 
(1) while Python can directly employ the `\texttt{sort}' function, C++ has to load it from the `\texttt{algorithm}' library. Regarding variables, Python's `\texttt{list}' is more similar to C++'s `\texttt{vector}' than its array. (2) List comprehension like `\texttt{sum(1 for penguin in penguins if penguin["age"] < 8)}' is a standard syntax in Python. However, a straightforward for-loop is the common practice in other PLs. (3) Static PLs such as C++ require to define the variable type. We carefully define `\texttt{int}' and `\texttt{double}' variables to ensure computational accuracy and enhance flexibility by defining `\texttt{struct}'. (4) We keep the naming styles of each PL. For instance, Python uses Snake Case, whereas Java favors Camel Case (`\texttt{secondPenguin}'). Appendix~\ref{sec:prompt} shows the demonstrations. 
The above examples present the variations in example PoTs across different PLs. To accurately assess the model's capability in a specific PL, it is crucial to carefully consider its characteristics during the process of constructing.

Based on identical reasoning process, we successfully craft demonstrations of each PL exhibiting its characteristics.
By adding the question after the demonstration, we get the prompt for each PL.

\subsection{Integration}
\label{sec:intergrate}
While Self-Consistency enhances performance by sampling to explore more reasoning paths, it can lead to repeated errors across different samples. In contrast, MultiPoT constructs multilingual prompts and generates PoTs in multiple PLs, significantly increasing the diversity of results. 

Specifically, after constructing prompts for each PL, models generate corresponding PoTs, while tracking cumulative probabilities. These probabilities indicate the model's confidence in each answer, with higher probabilities denoting greater confidence. PoTs are then executed and results are collected. The final answer is determined by voting on these results. In cases of tied votes, answers with higher cumulative probabilities are favored. The integration of multiple PLs introduces more potential correct answers and reduces the probability of the same errors in candidate results.

\section{Experiment Setup}

\subsection{Programming Languages} 

When selecting PLs to compare with Python, we focus on diversity. 
JavaScript is the most popular language on GitHub~\cite{github2023state} and has less overlap in application with Python, particularly in the ML/AI domains. 
R is a flexible and powerful language like Python but has much less data in pre-training data.
The three PLs above are dynamic languages that do not require explicit variable type definitions. 
To incorporate the diversity of language types, we select the two most common static languages, Java and C++. The latter is closer to low-level programming and has fewer extension packages. 
We do not include C due to its high similarity with C++.
These five languages offer a diverse range of application scenarios, data volumes, and language types compared to Python.

\begin{figure*}[!ht]
\centering
\includegraphics[width=\textwidth]{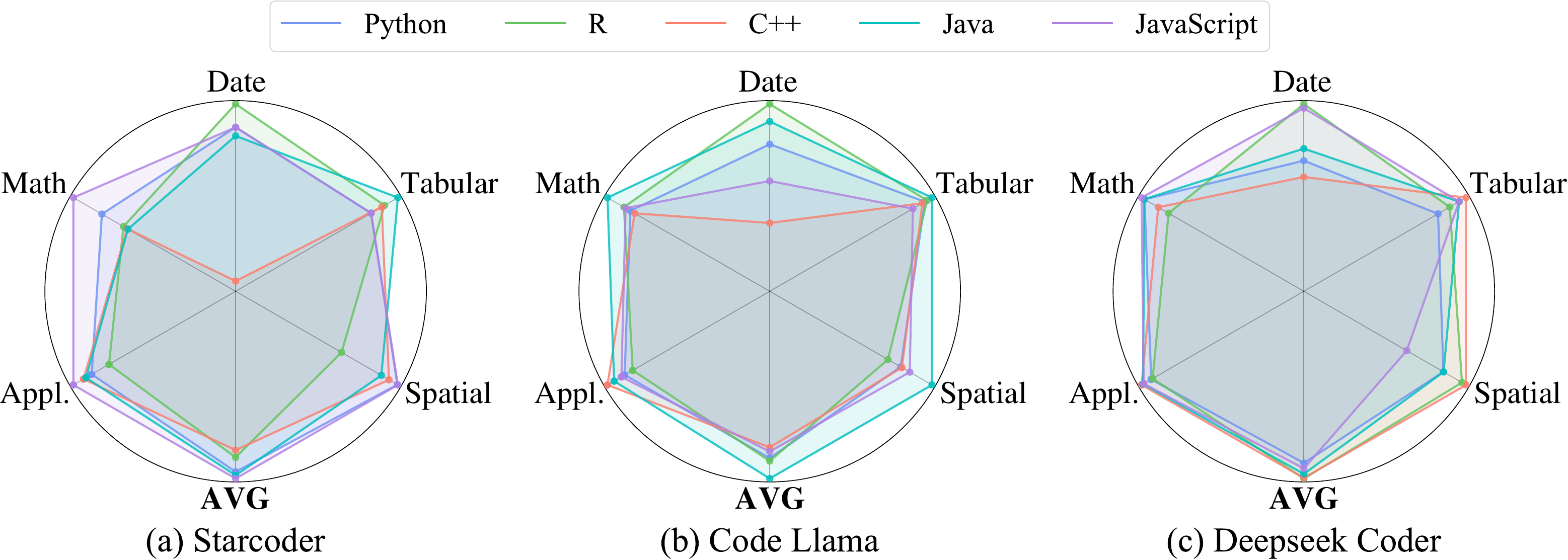}
\caption{The greedy decoding performance of three models across five tasks in five different PLs. AVG denotes the average performance of a PL across all tasks. Each language performance is expressed as a ratio to the highest-performing language for that specific task. The center of the circle represents 50\%.
Detailed numerical data are provided in the Table~\ref{tab:re1} in Appendix ~\ref{sec:appendix_results}.
}
\label{fig:greedy_results}
\end{figure*}

\begin{table*}[!ht]
\centering
\begin{small}
\begin{tabular}{l|rrrrrr|rrrrrr}
\toprule
\multirow{2}{*}{Language}& \multicolumn{6}{c|}{Code LLMs} & \multicolumn{6}{c}{ChatGPT} \\ 
\cmidrule{2-13}
& Appl. & Math & Date & Tabular & Spatial & \textbf{AVG} & Appl. & Math & Date & Tabular & Spatial & \textbf{AVG} \\ 
\midrule
Python & 58.51 & 23.62 & 42.37 & 83.00 & 73.87 & 56.27 & 80.75 & 39.74 & 46.61 & \textbf{94.63} & 91.70 & 70.69\\ 
R & 57.04 & 22.61 & \textbf{47.70} & 85.46 & 71.20 & 56.80 & 79.37 & 34.86 & \textbf{55.01} & 89.93 & \textbf{92.85} & 70.40 \\ 
C++ & \textbf{60.80} & 22.61 & 32.79 & 86.35 & \textbf{75.87} & 55.68 & 79.46 & 39.90 & 47.70 & 91.95 & 86.65 & 69.13 \\ 
Java & 60.11 & 23.75 & 43.81 & \textbf{87.92} & 75.82 & \textbf{58.28} & 80.63 & \textbf{42.65} & 51.22 & 87.92 & 86.70 & 69.82 \\ 
JavaScript & 60.14 & \textbf{24.35} & 42.82 & 83.89 & 71.58 & 56.56 & \textbf{81.25} & 36.07 & \textbf{55.01} & 92.62 & 90.15 & \textbf{71.02} \\ 
\bottomrule
\end{tabular}%
\end{small}
\caption{The performance of Code LLMs and ChatGPT for greedy decoding for five languages on five tasks. Code LLMs are the average results for Starcoder, Code Llama, and Deepseek Coder. \textbf{AVG} means the average performance of the language on five tasks. \textbf{Bold} denotes the highest performance on the task.}
\label{tab:greedy_results}
\end{table*}

\subsection{Tasks}

We select representative and discriminating tasks.
We initially select four tasks from ~\citet{Pal}: \textbf{Math Application (Appl.)}, \textbf{Date}, \textbf{Tabular} and \textbf{Spatial}, and add the task \textbf{Math}.
Appl. contains elementary-level mathematical application problems (GSM8K, SVAMP, Asdiv~\citep{gsm8k, SVAMP, ASDIV}). Date, Tabular, and Spatial are extracted from BBH-Hard~\cite{BBH-hard} (Date Understanding, Penguins in a Table, Reasoning about Coloured Objects). These tasks assess understanding and reasoning about temporal sequences, structured text, and spatial positioning respectively. Math, consisting of the transformed MATH~\citep{MATH} dataset. The difference between Math and Appl. lies in the level of difficulty. Math is more challenging and directly describes the math question without scenarios.
The five tasks are distinct and representative of the evaluation of reasoning capabilities. They are language-agnostic, meaning that they can be performed in any PL, effectively demonstrating the model's reasoning ability across different languages.
The additional details of the tasks are in the Appendix~\ref{sec:tasks}.

\subsection{Metric}
Remaining consistent with previous work~\cite{PoT, Pal}, the metric is accuracy. For tasks whose ground truth are real numbers (Appl./Math), the answer is considered correct if its difference from the ground truth is less than 1e-3; for tasks with string-type ground truth (Date/Tabular/Spatial), the answer is considered correct only if it is exactly the same as the ground truth.

\subsection{Backbone LLMs}
As the previously used code-davinci family is no longer accessible, we select four backbone LLMs, including the three strongest Code LLMs: \textbf{Starcoder} (15B), \textbf{Code Llama} (34B), and \textbf{Deepseek Coder} (33B). 
We select the base versions.
The experiments of the Python version are discussed in Section~\ref{python_model}, and the results are consistent with our conclusions and methodology. 
\textbf{ChatGPT} is also utilized as a representative of code-capable NL LLMs, invoking through the API of gpt-3.5-turbo-0701.
By choosing these backbone LLMs with different sizes and characteristics, we can obtain more realistic and credible results.

\subsection{Inference Details} 
We combine ~\citet{PoT} and ~\citet{Pal}'s prompt templates for few-shot inference. We fix the questions from the previous work and write code in the respective PLs.
The number of questions in each task is shown in Appendix~\ref{sec:tasks}.
When sampling for Self-Consistency, we follow \citet{PoT} and set $t=0.4, top\_p=1$. For a fair comparison with MultiPoT which integrates five languages, we set $k=5$.

\section{Results}
\begin{table*}[t]
\centering
\begin{small}
\begin{tabular}{l|rrrrrr|rrrrrr}
\toprule
& \multicolumn{6}{c|}{ChatGPT} & \multicolumn{6}{c}{Starcoder} \\
\cmidrule{2-13}
& Appl. & Math & Date & Table & Spatial & \textbf{AVG} & Appl. & Math & Date & Table & Spatial & \textbf{AVG} \\
\midrule
    Python & 82.31 & 45.76 & 47.70 & 94.63 & 93.60 & 72.80 & 47.04 & 19.69 & 34.96 & 79.19 & 70.00 & 50.18 \\
    R & 80.95 & 40.61 & \textbf{58.81} & 93.29 & 94.60 & 73.65 & 44.21 & 17.74 & 37.13 & 77.85 & 65.90 & 48.57 \\
    C++ & 81.40 & 43.77 & 49.05 & 93.29 & 88.45 & 71.19 & 47.34 & 16.74 & 18.70 & 82.55 & 70.95 & 47.26 \\
    Java & 81.79 & 45.33 & 53.39 & 92.62 & 88.80 & 72.39 & 47.97 & 16.76 & 35.23 & 78.52 & 69.50 & 49.60 \\
    JavaScript & 82.58 & 40.64 & 56.10 & 96.64 & 93.30 & 73.85 & 48.40 & 19.15 & 36.31 & 80.54 & \textbf{72.95} & 51.47 \\
    \midrule
    \textbf{MultiPoT} & \textbf{84.33} & \textbf{49.92} & 58.54 & \textbf{98.66} & \textbf{95.30} & \textbf{77.35} & \textbf{49.67} & \textbf{20.41} & \textbf{40.38} & \textbf{87.25} & 71.55 & \textbf{53.85} \\
\midrule
& \multicolumn{6}{c|}{Code Llama} & \multicolumn{6}{c}{Deepseek Coder} \\
\midrule
    Python & 68.63 & 27.95 & 50.68 & 92.62 & 77.55 & 63.48 & 70.65 & \textbf{37.64} & 44.72 & 93.96 & 89.80 & 67.35 \\
    R & 66.80 & 26.65 & 58.27 & 93.29 & 79.05 & 64.81 & 69.22 & 33.59 & 53.12 & 93.29 & 92.60 & 68.36 \\
    C++ & \textbf{71.33} & 24.99 & 43.36 & 93.29 & 80.45 & 62.68 & \textbf{72.32} & 33.94 & 39.57 & \textbf{95.30} & \textbf{93.40} & 66.91 \\
    Java & 70.10 & 27.93 & 56.91 & \textbf{93.96} & \textbf{81.80} & 66.14 & 72.10 & 35.35 & \textbf{55.56} & 93.96 & 88.75 & 69.14 \\
    JavaScript & 68.97 & 26.16 & 50.41 & 87.25 & 80.35 & 62.63 & 71.89 & 35.60 & 52.57 & 93.29 & 86.10 & 67.89 \\
    \midrule
    \textbf{MultiPoT} & 71.17 & \textbf{27.97} & \textbf{58.54} & \textbf{93.96} & 79.60 & \textbf{66.24} & \textbf{72.32} & 37.55 & 54.47 & \textbf{95.30} & 91.70 & \textbf{70.27} \\
\bottomrule
\end{tabular}
\end{small}
\caption{Self-Consistency and MultiPoT results of four LLMs on five tasks and \textbf{AVG}.}
\label{tab:multipot_performance}
\end{table*}
In this section, we first discover that Python is not the best language for all tasks and all models from the results of greedy decoding. There is no such perfect language. The performance of each PL varies greatly depending on the task and model (Section~\ref{sec:greedy}). After Self-Consistency, the performance discrepancy still exists. Finally, by integrating multiple languages, MultiPoT significantly outperforms Python. Furthermore, its performance matches or exceeds the best monolingual PoTs in almost all scenarios and achieves improvement on task and model averages (Section~\ref{sec:multipot}).


\subsection{Comparison among PLs}
\label{sec:greedy}



\textbf{Python is not the optimal language choice.} 
Figure~\ref{fig:greedy_results} shows the performance gap between each language and the best-performing language on each task of the three Code LLMs.
It illustrates that Python does not achieve the best performance on any of the tasks for any of the Code LLMs. On Deepseek Coder, Python is even the worst on average. 
Table~\ref{tab:greedy_results} shows the greedy decoding results of ChatGPT.
Although Python performs best on Tabular, it falls short by 2.9\% and 8.4\% compared to the best PL on Math and Date respectively. 
The preference for Python among humans may be due to its simple syntax and high readability, but it is a subjective bias that PoT only needs it. Relying on Python leads to a suboptimal outcome. 

However, it is important to note that \textbf{there is no one-size-fits-all language}. The gap between PLs is significant when considering each task and model.

\textbf{The performance of each PL is task-dependent}.
\textit{AVG performance does not fully capture the disparity among languages}. Java and JavaScript performances of Starcoder differ by only 0.41\% on AVG, but by 6.71\% on Tabular. While the difference between the best and worst PLs of ChatGPT on AVG is less than 2\% in Table~\ref{tab:greedy_results}, there are four tasks whose gap among languages exceeds 6\%.
\textit{Different languages are suitable for different tasks.}
Table~\ref{tab:greedy_results} indicates that, except for C++, all PLs excel in at least one task on ChatGPT. 
Moreover, on ChatGPT, except for JavaScript, each language also ranks as the least effective in at least one task.
\textit{A language that performs exceptionally well in one task might underperform in another.} For instance, R demonstrates superior performance on Date for both Code LLMs and ChatGPT, yet it is the least effective on Appl. and Math. 

\textbf{The performance of each PL is model-dependent.}
\textit{Code LLMs and ChatGPT differ significantly.}
The results of three Code LLMs are averaged and compared with ChatGPT in Table~\ref{tab:greedy_results}.
It shows that, on Appl., C++ performs best on Code LLMs but ranks second-to-last on ChatGPT; on Math, JavaScript excels on Code LLMs but similarly ranks second-to-last on ChatGPT; and on Spatial, Java ranks second-highest on Code LLMs (with only a 0.05\% less than C++) but is second-to-last on ChatGPT.
\textit{Even within Code LLMs, disparities between models are evident. }
Figure~\ref{fig:greedy_results} shows that Code Llama has a clear preference for Java, which keeps the top two ranks across all tasks, yet is not observed on the remaining models. 
On Deepseek Coder, C++ leads on average, whereas it ranks last on the other models. R ranks second on Spatial on Deepseek Coder, but the worst on the other two Code LLMs.
These variations demonstrate that \textbf{different PLs exhibit unique strengths and diversity} due to complex factors such as task suitability and model preference. 
A further error analysis of the experimental results is shown in Appendix~\ref{sec:error_analyse}.

\subsection{Comparision between Self-Consistency and MultiPoT}
\label{sec:multipot}
\textbf{Self-Consistency does not eliminate performance disparities between PLs}, despite it significantly improving the performance. Table~\ref{tab:multipot_performance} presents the Self-Consistency results. \textit{The inherent strength of different languages persist.} The optimal PL on each scenario is generally consistent with greedy decoding results, except Python emerges as the superior language on Math on all model. \textit{The weaknesses of each language is further amplified}. For example, on Date of Deepseek Coder, C++ already had the lowest performance in greedy decoding, and Self-Consistency increases this gap even more. As a result, C++ shifts from the highest average performance in greedy decoding on Deepseek Coder to the lowest in Self-Consistency, despite remaining the best on Appl., Tabular, and Spatial.
\textit{A single language offers limited diversity.} When faced with tasks outside its strength, monolingual samples often make the same mistakes repeatedly, resulting in incorrect answers being chosen through voting.


Different from Self-Consistency relying on a single PL, \textbf{MutliPoT} integrates multiple PLs. It not only \textbf{leverages the distinct strength} of each PL, 
but also \textbf{utilizes their greater diversity} to reduce the probability of repeating the same errors.  

\textbf{MultiPoT significantly outperforms Python on almost all scenarios}. \textit{It enhances performance in tasks or models where Python is weak.} Across the four models, MultiPoT improves upon Python's performance on Date by at least 15\%, and in average (AVG) performance by 4.33\% to 7.32\%. Furthermore, \textit{MultiPoT also capitalizes on Python's strength.} On Math, where Python excels, MultiPoT also achieves the best results, except in Deepseek Coder, where it slightly trails Python but remains significantly ahead of other languages.

\textbf{MultiPoT achieves comparable or superior performance to the best monolingual results across all tasks and models.}  
\textit{It is task-agnostic}. It surpasses Self-Consistency on four tasks, ranking second 
on the remaining task, regardless of whether on Code LLMs average 
(Table~\ref{tab:languages_avg_sc_results})
or ChatGPT. \textit{MultiPoT is also model-agnostic}. It is the top performer across all LLMs on Tabular. On AVG, MultiPoT outperforms the best monolingual result on all four models. Particularly on ChatGPT and Starcoder, it exhibits an improvement of over 4.6\%.

The performance of PLs depends on the task and model. Analyzing the interplay of PL, task, and model in practical applications is challenging. Therefore, MultiPoT is a great choice which has consistently high performance across scenarios.

\begin{figure}[t]
\centering
\includegraphics[width=\columnwidth]{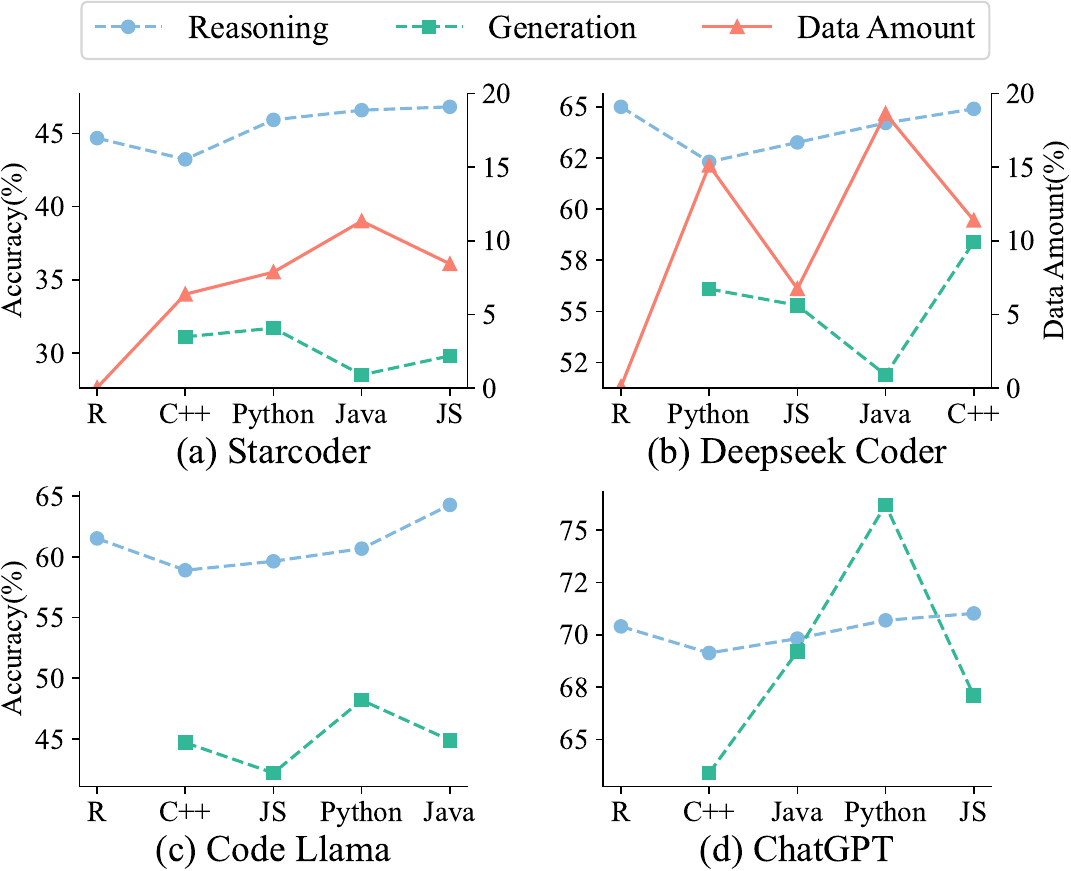}
\caption{The reasoning ability, code generation ability, and percentage in pre-training data for different languages. Generation lacks data for R. The horizontal coordinates of each model are ranked according to the rise in reasoning performance (excluding R).
}
\label{fig:factors}
\end{figure} 


\section{Discussion}
\subsection{Reasoning Ability of Different Languages}
In Section~\ref{sec:greedy}, we note that the ranking of the average performance of PL varies on each model.
The language distribution in the pre-training data of Starcoder and Deepseek Coder offers insights into whether data amount, defined as the percentage of each language in the pre-training corpus, impacts reasoning capabilities. Moreover, 
we are interested in examining whether code generation and reasoning of multilingual ability are aligned. The difference between the two tasks is elucidated in Appendix~\ref{sec:task_difference}. To assess code generation ability, we utilize the results of each model on the Multilingual HumanEval benchmark, focusing on the four available languages, excluding R due to a lack of evaluation dataset.
\textbf{Data distribution influences but does not completely determine reasoning ability.} Figure~\ref{fig:factors} shows the relative relationships among reasoning performance of C++, Python, and Java are consistent with data distribution on Starcoder. However, R demonstrates unexpectedly strong performance, which has an extremely low percentage in both models. C++ has less data amount than Java on Deepseek Coder, but better reasoning performance. This suggests that there are other factors affecting performance besides data distribution.

\textbf{Code generation abilities do not always align with reasoning abilities.} We compare the four languages excluding R in Figure~\ref{fig:factors}. On ChatGPT, the reasoning and code generation abilities of C++, Java, and Python align perfectly. However, an opposite trend is observed in Deepseek Coder’s Python, JavaScript, and Java, where the two abilities diverge significantly. It highlights the necessity of testing the reasoning abilities of different PLs.
\begin{table*}[ht]
\centering
\begin{small}
\begin{tabular}{l|ccc|ccc}
\toprule
\#Shots & \multicolumn{3}{c|}{0} & \multicolumn{3}{c}{3} \\
\midrule
~ & RE & WA & AC & RE & WA & AC \\
\midrule
Python    & 17.33 & 29.97 & \textbf{52.71} & \textbf{2.72} & 32.14 & 65.14 \\
R       & 55.81 & 15.45 & 28.74 & 2.08 & \textbf{34.47} & 63.44 \\
C++         & 59.61 & 14.90 & 25.48 & 0.34 & 30.87 & \textbf{68.79} \\
Java      & 14.86 & \textbf{33.25} & 51.89 & 0.54 & 32.07 & 67.38 \\
JavaScript & \textbf{69.26} & 13.77 & 16.96 & 1.74 & 32.41 & 65.84 \\
\bottomrule
\end{tabular}
\end{small}
\caption{Comparison for different PLs under zero-shot (0) and 3-shot (3) demonstrations. \textbf{RE} represents Runtime Error. \textbf{WA} means Wrong Answer. \textbf{AC} represents ACcuracy.}
\label{tab:zero-shot}
\end{table*}

\begin{table}[t]
\centering
\begin{small}
\begin{tabular}{lcccc}
\toprule
 & StarC.  & C. Llama & Deep.C. & GPT \\ 
 \midrule
Python  & 61.03  & 73.23 & 75.80 & 77.62 \\
R       & 58.86  & 75.11 & 76.02 & 79.00 \\
C++     & 59.75  & 72.82 & 75.80 & 77.82 \\
Java    & 61.32  & 75.62 & 78.06 & 78.08 \\
JavaScript & 62.60  & 74.15 & 76.62 & 77.65 \\
\midrule
MultiPoT   & \textbf{64.52}  & \textbf{75.71}       & \textbf{78.41}       & \textbf{83.94}  \\ 
\bottomrule
\end{tabular}
\end{small}
\caption{The average coverage rate on five tasks of Self-Consistency and MultiPoT on each model.}
\label{tab:oracle}
\end{table}

\begin{table}[t]
\centering
\begin{small}
\begin{tabular}{lcc}
\toprule
\textbf{Stability Metric} & \textbf{Starcoder} & \textbf{Deepseek Coder}  \\ 
\midrule
Default                & \textbf{53.85}   & \textbf{70.27}             \\
Length Short          & 53.36   & 69.99                                \\
Length Long           & 53.16  & 69.76                                 \\
Random                & 53.71     & 69.99                                \\
Data Amount Little    & 53.18     & 70.20                                \\
Data Amount Large      & 53.55    & 69.43                               \\
\midrule
$\Delta$                 & 0.69      & 0.84                                 \\ 
\bottomrule
\end{tabular}
\end{small}
\caption{The performance of MultiPoT with different sorting methods. Length Short/Long represents the ascending/descending order according to the length of PoTs, respectively. $\Delta$ denotes the range of change.}
\label{tab:stability}
\end{table}


\textbf{Zero-shot reasoning ability shows considerable inconsistency when compared to 3-shot reasoning ability.} Table~\ref{tab:zero-shot} presents the results of zero-shot and 3-shot experiments using Code Llama 34B on Appl., where AC denotes accuracy, and incorrect outcomes are further classified into Runtime Errors (RE) and Wrong Answers (WA). The results reveal particularly steep declines in R, C++, and JavaScript, largely driven by a significant increase in RE. This suggests that different PLs exhibit varying levels of sensitivity to shot settings. Two prominent error patterns emerge from the zero-shot outputs: (1) LLM frequently generates repetitive comments until reaching the maximum sequence length, and (2) LLM generates CoT without corresponding executable code. These observations highlight the importance of high-quality, language-specific demonstrations; only with effective demonstrations can the model fully harness the reasoning capabilities of different PLs.

\subsection{MultiPoT Analysis}
\begin{table*}[t]
\centering
\begin{small}
\begin{tabular}{llcccccc}
\toprule
Model & Method & Appl. & Math & Date & Table & Spatial & AVG \\ 
\midrule
\multirow{2}{*}{Base} & Python & 68.63 & 27.95 & 50.68 & 92.62 & 77.55 & 63.48 \\
& MultiPoT & \textbf{71.17} & 27.95 & \textbf{58.54} & \textbf{93.96} & \textbf{79.60} & \textbf{66.24} \\ 
\midrule
\multirow{2}{*}{Python} & Python & 69.54 & \textbf{28.46} & 48.24 & 91.28 & 74.65 & 62.43 \\ 
 & MultiPoT & 70.67 & 27.46 & 55.83 & 92.62 & 76.70 & 64.65 \\ 
 \bottomrule
\end{tabular}
\end{small}
\caption{The performance of Python Self-Consistency and MultiPoT on Code Llama Base and Code Llama Python.}
\label{tab:code_llama_python}
\end{table*}

\textbf{MutliPoT has the highest coverage rate}.
\label{sec:oracle}
Unlike the voting mechanism which requires a majority for the correct answer, the coverage rate 
is the percentage of questions that have at least one correct answer in five candidate answers in the dataset. For example, if the candidate answers to a question are ``(6 5 5 5 7)'' and the ground truth is ``7'', the question is covered. Coverage rate can be considered as an upper bound because this metric represents the proportion of all potentially solvable problems, assuming there exists a selection mechanism better than the current voting mechanism.
Table~\ref{tab:oracle} demonstrates coverage rates on all four models and MultiPoT achieves the highest. 
The monolingual sampling covers less than the multilingual attempts, highlighting that the strength of different PLs exists. MultiPoT effectively utilizes the strength of different PLs and has the highest upper bound.

\textbf{MutliPoT has stable performance}. 
When results are tied, the top-ranked result is selected. Different sorting methods reflect the stability. Table~\ref{tab:stability} shows the performance fluctuation. MultiPoT is less than 1\% across various sorting criteria, including PoT length, randomness, or data amount from pre-training, compared to the default cumulative probability sorting. This indicates that MultiPoT consistently selects the correct answer directly, with few instances of ties with incorrect answers. This also suggests a lower probability of different PoTs making the same errors.

\begin{figure}[t]
\centering
\includegraphics[width=\columnwidth]{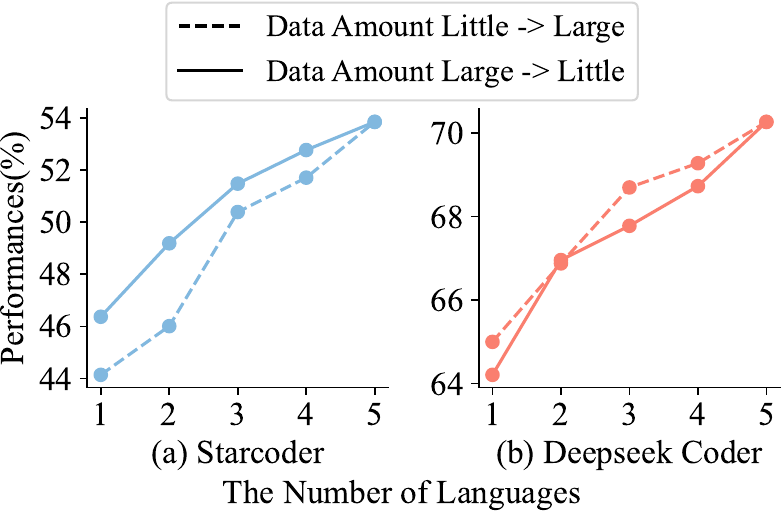}
\caption{The impact of the number of integrating PLs. We test the different order of adding languages.}
\label{fig:language_num}
\end{figure}
\textbf{More PLs are better}.
We investigate the impact of the number of PLs on MultiPoT. On both Starcoder and Deepseek Coder, we incrementally add languages in both ascending and descending order of data amount in Figure~\ref{fig:language_num}. The results show that MultiPoT's performance improves with more PLs, regardless of the order. This suggests that MultiPoT is highly scalable and performance can be further enhanced by incorporating more PLs. 


\begin{table}[t]
\centering
\begin{small}
\begin{tabular}{lcc}
\toprule
\textbf{Type} & \textbf{Starcoder} & \textbf{ChatGPT}  \\ 
\midrule
All Dynamic              & 50.41   & 74.92             \\
Dynamic + Static         & \textbf{51.87}   & \textbf{75.77}             \\
\bottomrule
\end{tabular}
\end{small}
\caption{The impact of different language type combinations on MultiPoT. All Dynamic indicates that the three languages are all dynamic, and Dynamic+Static indicates a combination of dynamic and static languages.}
\label{tab:languages_type}
\end{table}

\textbf{More language types are better}.
Python, R, and JavaScript are dynamic languages, while C++ and Java are static. To investigate whether a diverse set of language types enhances MultiPoT's performance, we focus on three PLs. On Starcoder and ChatGPT, JavaScript emerges as the highest-performing dynamic language, surpassing Java, which leads between the static languages. Consequently, we integrate JavaScript, Python, and R as All Dynamic and combine Java, Python, and R to represent Dynamic + Static. The results in Table~\ref{tab:languages_type} indicate that replacing the higher-performing JavaScript with the lower-performing Java improves performance. This suggests that more language types can provide more diversity to MultiPoT, thereby further enhancing performance.


\textbf{MultiPoT also works on Python model}.
\label{python_model}
Our prior experiments with Code LLMs utilize the Base version. However, Code LLMs also have a Python-specific version trained with additional Python corpora. Evaluating MultiPoT on this Python version, as shown in Table~\ref{tab:code_llama_python}, we find that Python Self-Consistency improves on Appl. and Math but declines on the other tasks compared to the Base model. Moreover, MultiPoT still outperforms Python Self-Consistency on all tasks except Math, highlighting the adaptability of MultiPoT. Notably, MultiPoT's performance on the Python model is lower across all tasks than on the Base model. This suggests that extensive training on monolingual corpora might diminish the Base model's multilingual abilities on reasoning tasks.

\begin{table}[t]
\centering
\begin{small}
\begin{tabular}{lccc}
\toprule
~ & Date & Table & Spatial \\ 
\midrule
		
CoT              & \textbf{82.38}   & 91.28   & 97.70 \\
PoT           & 79.40   & 97.28   & 97.95 \\
MultiPoT         & 80.22   & \textbf{98.66}   & \textbf{99.10} \\
\bottomrule
\end{tabular}
\end{small}
\caption{Comparison between Self-Consistency of CoT, PoT (Python), and MultiPoT. CoT results are based on Deepseek LLM v2, while PoT and MultiPoT are based on Deepseek Coder v2.}
\label{tab:cot}
\end{table}

\textbf{MultiPoT is better than CoT Self-Consistency.}
To compare the performance of CoT and PoT in scenarios where precise mathematical calculations are not required, we conduct experiments on Date, Table, and Spatial—using Deepseek LLM v2 (a 405B MoE LLM) for CoT and Deepseek Coder v2 (continually pretrained from Deepseek LLM v2) for PoT. The results, shown in Table~\ref{tab:cot}, indicate that PoT achieves better Self-Consistency than CoT on Table and Spatial, with MultiPoT further improving performance. On Table, the improvement demonstrates the advantage of PoT in understanding and reasoning over structured data. On Date, however, PoT slightly underperforms CoT, which is primarily due to PoT interpreting the difference between two dates as exclusive, while natural language typically considers it inclusive. Nevertheless, the results suggest that PoT remains valuable in scenarios where precise calculations are unnecessary, and MultiPoT continues to be effective.

\section{Conclusion}
Regarding the reliance on Python in PoT, we conducted extensive experiments across various models and tasks using multiple PLs. Our findings show that Python is not always the best choice; the optimal language depends on the specific task and model. Building on this insight, we introduce MultiPoT, a simple yet effective multilingual integrated method that leverages the strengths and diversity of different PLs. MultiPoT significantly outperforms Python and achieves matches or exceeds performance to the best monolingual outcomes in nearly all scenarios. With its high stability, MultiPoT offers a promising avenue for future research.
\clearpage
\section*{Limitations}
Our study is comprehensive, but has certain limitations that we plan to address in future research. Due to computational resource constraints, we confine our experiments to a select number of commonly used programming languages (PLs). While these PLs are representative, they do not encompass the entire spectrum of languages used in programming. Future research could investigate the advantages of incorporating a broader range of programming languages. This may reveal further insights and improve the relevance of our findings.

\section*{Ethical Considerations}
Our research utilizes publicly available models and datasets with proper citations and adheres to the usage guidelines of ChatGPT, minimizing the risk of generating toxic content due to the widely-used, non-toxic nature of our datasets and prompts.

\section*{Acknowledge}
We gratefully acknowledge the support of the National Natural Science Foundation of China (NSFC) via grant 62236004, 62206078, 62441603 and 62476073 and the support of Du Xiaoman (Beijing) Science Technology Co., Ltd.

\bibliography{custom}

\begin{thebibliography}{39}
\expandafter\ifx\csname natexlab\endcsname\relax\def\natexlab#1{#1}\fi

\bibitem[{Brown et~al.(2020)Brown, Mann, Ryder, Subbiah, Kaplan, Dhariwal, Neelakantan, Shyam, Sastry, Askell, Agarwal, Herbert-Voss, Krueger, Henighan, Child, Ramesh, Ziegler, Wu, Winter, Hesse, Chen, Sigler, Litwin, Gray, Chess, Clark, Berner, McCandlish, Radford, Sutskever, and Amodei}]{gpt3}
Tom Brown, Benjamin Mann, Nick Ryder, Melanie Subbiah, Jared~D Kaplan, Prafulla Dhariwal, Arvind Neelakantan, Pranav Shyam, Girish Sastry, Amanda Askell, Sandhini Agarwal, Ariel Herbert-Voss, Gretchen Krueger, Tom Henighan, Rewon Child, Aditya Ramesh, Daniel Ziegler, Jeffrey Wu, Clemens Winter, Chris Hesse, Mark Chen, Eric Sigler, Mateusz Litwin, Scott Gray, Benjamin Chess, Jack Clark, Christopher Berner, Sam McCandlish, Alec Radford, Ilya Sutskever, and Dario Amodei. 2020.
\newblock \href {https://proceedings.neurips.cc/paper_files/paper/2020/file/1457c0d6bfcb4967418bfb8ac142f64a-Paper.pdf} {Language models are few-shot learners}.
\newblock In \emph{Advances in Neural Information Processing Systems}, volume~33, pages 1877--1901. Curran Associates, Inc.

\bibitem[{Chen et~al.(2021)Chen, Tworek, Jun, Yuan, de~Oliveira~Pinto, Kaplan, Edwards, Burda, Joseph, Brockman, Ray, Puri, Krueger, Petrov, Khlaaf, Sastry, Mishkin, Chan, Gray, Ryder, Pavlov, Power, Kaiser, Bavarian, Winter, Tillet, Such, Cummings, Plappert, Chantzis, Barnes, Herbert-Voss, Guss, Nichol, Paino, Tezak, Tang, Babuschkin, Balaji, Jain, Saunders, Hesse, Carr, Leike, Achiam, Misra, Morikawa, Radford, Knight, Brundage, Murati, Mayer, Welinder, McGrew, Amodei, McCandlish, Sutskever, and Zaremba}]{codex}
Mark Chen, Jerry Tworek, Heewoo Jun, Qiming Yuan, Henrique~Ponde de~Oliveira~Pinto, Jared Kaplan, Harri Edwards, Yuri Burda, Nicholas Joseph, Greg Brockman, Alex Ray, Raul Puri, Gretchen Krueger, Michael Petrov, Heidy Khlaaf, Girish Sastry, Pamela Mishkin, Brooke Chan, Scott Gray, Nick Ryder, Mikhail Pavlov, Alethea Power, Lukasz Kaiser, Mohammad Bavarian, Clemens Winter, Philippe Tillet, Felipe~Petroski Such, Dave Cummings, Matthias Plappert, Fotios Chantzis, Elizabeth Barnes, Ariel Herbert-Voss, William~Hebgen Guss, Alex Nichol, Alex Paino, Nikolas Tezak, Jie Tang, Igor Babuschkin, Suchir Balaji, Shantanu Jain, William Saunders, Christopher Hesse, Andrew~N. Carr, Jan Leike, Josh Achiam, Vedant Misra, Evan Morikawa, Alec Radford, Matthew Knight, Miles Brundage, Mira Murati, Katie Mayer, Peter Welinder, Bob McGrew, Dario Amodei, Sam McCandlish, Ilya Sutskever, and Wojciech Zaremba. 2021.
\newblock \href {http://arxiv.org/abs/2107.03374} {Evaluating large language models trained on code}.

\bibitem[{Chen et~al.(2022)Chen, Ma, Wang, and Cohen}]{PoT}
Wenhu Chen, Xueguang Ma, Xinyi Wang, and William~W. Cohen. 2022.
\newblock \href {https://doi.org/10.48550/ARXIV.2211.12588} {Program of thoughts prompting: Disentangling computation from reasoning for numerical reasoning tasks}.
\newblock \emph{CoRR}, abs/2211.12588.

\bibitem[{Chowdhery et~al.(2023)Chowdhery, Narang, Devlin, Bosma, Mishra, Roberts, Barham, Chung, Sutton, Gehrmann, Schuh, Shi, Tsvyashchenko, Maynez, Rao, Barnes, Tay, Shazeer, Prabhakaran, Reif, Du, Hutchinson, Pope, Bradbury, Austin, Isard, Gur-Ari, Yin, Duke, Levskaya, Ghemawat, Dev, Michalewski, Garcia, Misra, Robinson, Fedus, Zhou, Ippolito, Luan, Lim, Zoph, Spiridonov, Sepassi, Dohan, Agrawal, Omernick, Dai, Pillai, Pellat, Lewkowycz, Moreira, Child, Polozov, Lee, Zhou, Wang, Saeta, Diaz, Firat, Catasta, Wei, Meier-Hellstern, Eck, Dean, Petrov, and Fiedel}]{palm}
Aakanksha Chowdhery, Sharan Narang, Jacob Devlin, Maarten Bosma, Gaurav Mishra, Adam Roberts, Paul Barham, Hyung~Won Chung, Charles Sutton, Sebastian Gehrmann, Parker Schuh, Kensen Shi, Sasha Tsvyashchenko, Joshua Maynez, Abhishek Rao, Parker Barnes, Yi~Tay, Noam Shazeer, Vinodkumar Prabhakaran, Emily Reif, Nan Du, Ben Hutchinson, Reiner Pope, James Bradbury, Jacob Austin, Michael Isard, Guy Gur-Ari, Pengcheng Yin, Toju Duke, Anselm Levskaya, Sanjay Ghemawat, Sunipa Dev, Henryk Michalewski, Xavier Garcia, Vedant Misra, Kevin Robinson, Liam Fedus, Denny Zhou, Daphne Ippolito, David Luan, Hyeontaek Lim, Barret Zoph, Alexander Spiridonov, Ryan Sepassi, David Dohan, Shivani Agrawal, Mark Omernick, Andrew~M. Dai, Thanumalayan~Sankaranarayana Pillai, Marie Pellat, Aitor Lewkowycz, Erica Moreira, Rewon Child, Oleksandr Polozov, Katherine Lee, Zongwei Zhou, Xuezhi Wang, Brennan Saeta, Mark Diaz, Orhan Firat, Michele Catasta, Jason Wei, Kathy Meier-Hellstern, Douglas Eck, Jeff Dean, Slav Petrov, and Noah Fiedel. 2023.
\newblock \href {http://jmlr.org/papers/v24/22-1144.html} {Palm: Scaling language modeling with pathways}.
\newblock \emph{Journal of Machine Learning Research}, 24(240):1--113.

\bibitem[{Cobbe et~al.(2021)Cobbe, Kosaraju, Bavarian, Chen, Jun, Kaiser, Plappert, Tworek, Hilton, Nakano et~al.}]{gsm8k}
Karl Cobbe, Vineet Kosaraju, Mohammad Bavarian, Mark Chen, Heewoo Jun, Lukasz Kaiser, Matthias Plappert, Jerry Tworek, Jacob Hilton, Reiichiro Nakano, et~al. 2021.
\newblock Training verifiers to solve math word problems.
\newblock \emph{arXiv preprint arXiv:2110.14168}.

\bibitem[{Gao et~al.(2020)Gao, Biderman, Black, Golding, Hoppe, Foster, Phang, He, Thite, Nabeshima, Presser, and Leahy}]{pile}
Leo Gao, Stella Biderman, Sid Black, Laurence Golding, Travis Hoppe, Charles Foster, Jason Phang, Horace He, Anish Thite, Noa Nabeshima, Shawn Presser, and Connor Leahy. 2020.
\newblock \href {http://arxiv.org/abs/2101.00027} {The pile: An 800gb dataset of diverse text for language modeling}.

\bibitem[{Gao et~al.(2023)Gao, Madaan, Zhou, Alon, Liu, Yang, Callan, and Neubig}]{Pal}
Luyu Gao, Aman Madaan, Shuyan Zhou, Uri Alon, Pengfei Liu, Yiming Yang, Jamie Callan, and Graham Neubig. 2023.
\newblock Pal: Program-aided language models.
\newblock In \emph{International Conference on Machine Learning}, pages 10764--10799. PMLR.

\bibitem[{Gimeno et~al.(2023)Gimeno, Altché, and Leblond}]{alphacode2}
Felix Gimeno, Florent Altché, and Rémi Leblond. 2023.
\newblock Alphacode 2 technical report.
\newblock Technical report, AlphaCode Team, Google DeepMind.

\bibitem[{GitHub(2023)}]{github2023state}
GitHub. 2023.
\newblock \href {https://github.blog/2023-11-08-the-state-of-open-source-and-ai} {The state of open source and ai}.

\bibitem[{Guo et~al.(2024)Guo, Zhu, Yang, Xie, Dong, Zhang, Chen, Bi, Wu, Li, Luo, Xiong, and Liang}]{deepseek-coder}
Daya Guo, Qihao Zhu, Dejian Yang, Zhenda Xie, Kai Dong, Wentao Zhang, Guanting Chen, Xiao Bi, Y.~Wu, Y.~K. Li, Fuli Luo, Yingfei Xiong, and Wenfeng Liang. 2024.
\newblock \href {http://arxiv.org/abs/2401.14196} {Deepseek-coder: When the large language model meets programming -- the rise of code intelligence}.

\bibitem[{Gupta and Kembhavi(2023)}]{visualprogramming}
Tanmay Gupta and Aniruddha Kembhavi. 2023.
\newblock Visual programming: Compositional visual reasoning without training.
\newblock In \emph{Proceedings of the IEEE/CVF Conference on Computer Vision and Pattern Recognition}, pages 14953--14962.

\bibitem[{Hendrycks et~al.(2021)Hendrycks, Burns, Kadavath, Arora, Basart, Tang, Song, and Steinhardt}]{MATH}
Dan Hendrycks, Collin Burns, Saurav Kadavath, Akul Arora, Steven Basart, Eric Tang, Dawn Song, and Jacob Steinhardt. 2021.
\newblock Measuring mathematical problem solving with the math dataset.
\newblock In \emph{Thirty-fifth Conference on Neural Information Processing Systems Datasets and Benchmarks Track (Round 2)}.

\bibitem[{Jin et~al.(2023)Jin, Shahriar, Tufano, Shi, Lu, Sundaresan, and Svyatkovskiy}]{jin2023inferfix}
Matthew Jin, Syed Shahriar, Michele Tufano, Xin Shi, Shuai Lu, Neel Sundaresan, and Alexey Svyatkovskiy. 2023.
\newblock \href {http://arxiv.org/abs/2303.07263} {Inferfix: End-to-end program repair with llms}.

\bibitem[{Joshi et~al.(2023)Joshi, Cambronero~Sanchez, Gulwani, Le, Verbruggen, and Radiček}]{Repair}
Harshit Joshi, José Cambronero~Sanchez, Sumit Gulwani, Vu~Le, Gust Verbruggen, and Ivan Radiček. 2023.
\newblock \href {https://doi.org/10.1609/aaai.v37i4.25642} {Repair is nearly generation: Multilingual program repair with llms}.
\newblock 37:5131--5140.

\bibitem[{Khare et~al.(2023)Khare, Dutta, Li, Solko-Breslin, Alur, and Naik}]{khare2023understanding}
Avishree Khare, Saikat Dutta, Ziyang Li, Alaia Solko-Breslin, Rajeev Alur, and Mayur Naik. 2023.
\newblock \href {http://arxiv.org/abs/2311.16169} {Understanding the effectiveness of large language models in detecting security vulnerabilities}.

\bibitem[{Kocetkov et~al.(2022)Kocetkov, Li, Allal, Li, Mou, Ferrandis, Jernite, Mitchell, Hughes, Wolf et~al.}]{Stack}
Denis Kocetkov, Raymond Li, Loubna~Ben Allal, Jia Li, Chenghao Mou, Carlos~Mu{\~n}oz Ferrandis, Yacine Jernite, Margaret Mitchell, Sean Hughes, Thomas Wolf, et~al. 2022.
\newblock The stack: 3 tb of permissively licensed source code.
\newblock \emph{arXiv preprint arXiv:2211.15533}.

\bibitem[{Koncel-Kedziorski et~al.(2023)Koncel-Kedziorski, Krumdick, Lai, Reddy, Lovering, and Tanner}]{bizbench}
Rik Koncel-Kedziorski, Michael Krumdick, Viet Lai, Varshini Reddy, Charles Lovering, and Chris Tanner. 2023.
\newblock Bizbench: A quantitative reasoning benchmark for business and finance.
\newblock \emph{arXiv preprint arXiv:2311.06602}.

\bibitem[{Li et~al.(2023{\natexlab{a}})Li, Liang, Xia, Zeng, Levine, Sadigh, Hausman, Chen, Fei-Fei, and brian ichter}]{CoC}
Chengshu Li, Jacky Liang, Fei Xia, Andy Zeng, Sergey Levine, Dorsa Sadigh, Karol Hausman, Xinyun Chen, Li~Fei-Fei, and brian ichter. 2023{\natexlab{a}}.
\newblock \href {https://openreview.net/forum?id=tlRUbI0Yf3} {Chain of code: Reasoning with a language model-augmented code interpreter}.
\newblock In \emph{NeurIPS 2023 Foundation Models for Decision Making Workshop}.

\bibitem[{Li et~al.(2023{\natexlab{b}})Li, Allal, Zi, Muennighoff, Kocetkov, Mou, Marone, Akiki, Li, Chim et~al.}]{Starcoder}
Raymond Li, Loubna~Ben Allal, Yangtian Zi, Niklas Muennighoff, Denis Kocetkov, Chenghao Mou, Marc Marone, Christopher Akiki, Jia Li, Jenny Chim, et~al. 2023{\natexlab{b}}.
\newblock Starcoder: may the source be with you!
\newblock \emph{arXiv preprint arXiv:2305.06161}.

\bibitem[{Madaan and Yazdanbakhsh(2022)}]{madaan2022text}
Aman Madaan and Amir Yazdanbakhsh. 2022.
\newblock Text and patterns: For effective chain of thought, it takes two to tango.
\newblock \emph{arXiv preprint arXiv:2209.07686}.

\bibitem[{Miao et~al.(2020)Miao, Liang, and Su}]{ASDIV}
Shen-Yun Miao, Chao-Chun Liang, and Keh-Yih Su. 2020.
\newblock A diverse corpus for evaluating and developing english math word problem solvers.
\newblock In \emph{Proceedings of the 58th Annual Meeting of the Association for Computational Linguistics}, pages 975--984.

\bibitem[{Nguyen et~al.(2023)Nguyen, Nam, Dau, Nguyen, Nghiem, Guo, and Bui}]{vault}
Dung Nguyen, Le~Nam, Anh Dau, Anh Nguyen, Khanh Nghiem, Jin Guo, and Nghi Bui. 2023.
\newblock \href {https://doi.org/10.18653/v1/2023.findings-emnlp.316} {The vault: A comprehensive multilingual dataset for advancing code understanding and generation}.
\newblock In \emph{Findings of the Association for Computational Linguistics: EMNLP 2023}, pages 4763--4788, Singapore. Association for Computational Linguistics.

\bibitem[{Nijkamp et~al.(2023)Nijkamp, Pang, Hayashi, Tu, Wang, Zhou, Savarese, and Xiong}]{codegen}
Erik Nijkamp, Bo~Pang, Hiroaki Hayashi, Lifu Tu, Huan Wang, Yingbo Zhou, Silvio Savarese, and Caiming Xiong. 2023.
\newblock \href {http://arxiv.org/abs/2203.13474} {Codegen: An open large language model for code with multi-turn program synthesis}.

\bibitem[{Patel et~al.(2021)Patel, Bhattamishra, and Goyal}]{SVAMP}
Arkil Patel, Satwik Bhattamishra, and Navin Goyal. 2021.
\newblock Are nlp models really able to solve simple math word problems?
\newblock In \emph{Proceedings of the 2021 Conference of the North American Chapter of the Association for Computational Linguistics: Human Language Technologies}, pages 2080--2094.

\bibitem[{Roziere et~al.(2023)Roziere, Gehring, Gloeckle, Sootla, Gat, Tan, Adi, Liu, Remez, Rapin et~al.}]{code-llama}
Baptiste Roziere, Jonas Gehring, Fabian Gloeckle, Sten Sootla, Itai Gat, Xiaoqing~Ellen Tan, Yossi Adi, Jingyu Liu, Tal Remez, J{\'e}r{\'e}my Rapin, et~al. 2023.
\newblock Code llama: Open foundation models for code.
\newblock \emph{arXiv preprint arXiv:2308.12950}.

\bibitem[{Shypula et~al.(2023)Shypula, Madaan, Zeng, Alon, Gardner, Hashemi, Neubig, Ranganathan, Bastani, and Yazdanbakhsh}]{shypula2023learning}
Alexander Shypula, Aman Madaan, Yimeng Zeng, Uri Alon, Jacob Gardner, Milad Hashemi, Graham Neubig, Parthasarathy Ranganathan, Osbert Bastani, and Amir Yazdanbakhsh. 2023.
\newblock \href {http://arxiv.org/abs/2302.07867} {Learning performance-improving code edits}.

\bibitem[{Surís et~al.(2023)Surís, Menon, and Vondrick}]{vipergpt}
Dídac Surís, Sachit Menon, and Carl Vondrick. 2023.
\newblock \href {http://arxiv.org/abs/2303.08128} {Vipergpt: Visual inference via python execution for reasoning}.

\bibitem[{Suzgun et~al.(2022)Suzgun, Scales, Sch{\"a}rli, Gehrmann, Tay, Chung, Chowdhery, Le, Chi, Zhou et~al.}]{BBH-hard}
Mirac Suzgun, Nathan Scales, Nathanael Sch{\"a}rli, Sebastian Gehrmann, Yi~Tay, Hyung~Won Chung, Aakanksha Chowdhery, Quoc~V Le, Ed~H Chi, Denny Zhou, et~al. 2022.
\newblock Challenging big-bench tasks and whether chain-of-thought can solve them.
\newblock \emph{arXiv preprint arXiv:2210.09261}.

\bibitem[{Wang et~al.(2023{\natexlab{a}})Wang, Dou, Zhang, Zeng, and Che}]{wang2023exploring}
Dingzirui Wang, Longxu Dou, Wenbin Zhang, Junyu Zeng, and Wanxiang Che. 2023{\natexlab{a}}.
\newblock \href {http://arxiv.org/abs/2308.10585} {Exploring equation as a better intermediate meaning representation for numerical reasoning}.

\bibitem[{Wang et~al.(2024)Wang, Chen, Yuan, Zhang, Li, Peng, and Ji}]{codeact}
Xingyao Wang, Yangyi Chen, Lifan Yuan, Yizhe Zhang, Yunzhu Li, Hao Peng, and Heng Ji. 2024.
\newblock \href {http://arxiv.org/abs/2402.01030} {Executable code actions elicit better llm agents}.
\newblock In \emph{ICML}.

\bibitem[{Wang et~al.(2023{\natexlab{b}})Wang, Li, and Ji}]{wang-etal-2023-code4struct}
Xingyao Wang, Sha Li, and Heng Ji. 2023{\natexlab{b}}.
\newblock \href {https://doi.org/10.18653/v1/2023.acl-long.202} {{C}ode4{S}truct: Code generation for few-shot event structure prediction}.
\newblock In \emph{Proceedings of the 61st Annual Meeting of the Association for Computational Linguistics (Volume 1: Long Papers)}, pages 3640--3663, Toronto, Canada. Association for Computational Linguistics.

\bibitem[{Wang et~al.(2022)Wang, Wei, Schuurmans, Le, Chi, Narang, Chowdhery, and Zhou}]{self-consistency}
Xuezhi Wang, Jason Wei, Dale Schuurmans, Quoc~V Le, Ed~H Chi, Sharan Narang, Aakanksha Chowdhery, and Denny Zhou. 2022.
\newblock Self-consistency improves chain of thought reasoning in language models.
\newblock In \emph{The Eleventh International Conference on Learning Representations}.

\bibitem[{Wang et~al.(2023{\natexlab{c}})Wang, Wei, Schuurmans, Le, Chi, Narang, Chowdhery, and Zhou}]{wang2023selfconsistency}
Xuezhi Wang, Jason Wei, Dale Schuurmans, Quoc~V Le, Ed~H. Chi, Sharan Narang, Aakanksha Chowdhery, and Denny Zhou. 2023{\natexlab{c}}.
\newblock \href {https://openreview.net/forum?id=1PL1NIMMrw} {Self-consistency improves chain of thought reasoning in language models}.
\newblock In \emph{The Eleventh International Conference on Learning Representations}.

\bibitem[{Wei et~al.(2022)Wei, Wang, Schuurmans, Bosma, Xia, Chi, Le, Zhou et~al.}]{CoT}
Jason Wei, Xuezhi Wang, Dale Schuurmans, Maarten Bosma, Fei Xia, Ed~Chi, Quoc~V Le, Denny Zhou, et~al. 2022.
\newblock Chain-of-thought prompting elicits reasoning in large language models.
\newblock \emph{Advances in Neural Information Processing Systems}, 35:24824--24837.

\bibitem[{Wu et~al.(2023)Wu, Jiang, Pham, Lutellier, Davis, Tan, Babkin, and Shah}]{Wu_2023}
Yi~Wu, Nan Jiang, Hung~Viet Pham, Thibaud Lutellier, Jordan Davis, Lin Tan, Petr Babkin, and Sameena Shah. 2023.
\newblock \href {https://doi.org/10.1145/3597926.3598135} {How effective are neural networks for fixing security vulnerabilities}.
\newblock In \emph{Proceedings of the 32nd ACM SIGSOFT International Symposium on Software Testing and Analysis}, ISSTA ’23. ACM.

\bibitem[{Yang et~al.(2023)Yang, Wang, Lu, Liu, Le, Zhou, and Chen}]{yang2023large}
Chengrun Yang, Xuezhi Wang, Yifeng Lu, Hanxiao Liu, Quoc~V. Le, Denny Zhou, and Xinyun Chen. 2023.
\newblock \href {http://arxiv.org/abs/2309.03409} {Large language models as optimizers}.

\bibitem[{Yang et~al.(2024)Yang, Liu, Wu, Yang, Fung, Li, Huang, Cao, Wang, Wang, Ji, and Zhai}]{IA-Survey}
Ke~Yang, Jiateng Liu, John Wu, Chaoqi Yang, Yi~R. Fung, Sha Li, Zixuan Huang, Xu~Cao, Xingyao Wang, Yiquan Wang, Heng Ji, and Chengxiang Zhai. 2024.
\newblock \href {http://arxiv.org/abs/2401.00812} {If llm is the wizard, then code is the wand: A survey on how code empowers large language models to serve as intelligent agents}.

\bibitem[{Zhang et~al.(2023)Zhang, Nie, Li, and Gligoric}]{10.1145/3611643.3616350}
Jiyang Zhang, Pengyu Nie, Junyi~Jessy Li, and Milos Gligoric. 2023.
\newblock \href {https://doi.org/10.1145/3611643.3616350} {Multilingual code co-evolution using large language models}.
\newblock In \emph{Proceedings of the 31st ACM Joint European Software Engineering Conference and Symposium on the Foundations of Software Engineering}, ESEC/FSE 2023, page 695–707, New York, NY, USA. Association for Computing Machinery.

\bibitem[{Zhang et~al.(2024)Zhang, Hu, Xu, Yan, Xu, Jin, Zhang, and Huang}]{tinychart}
Liang Zhang, Anwen Hu, Haiyang Xu, Ming Yan, Yichen Xu, Qin Jin, Ji~Zhang, and Fei Huang. 2024.
\newblock Tinychart: Efficient chart understanding with visual token merging and program-of-thoughts learning.
\newblock \emph{arXiv preprint arXiv:2404.16635}.

\end{thebibliography}

\clearpage
\appendix
\renewcommand{\thesection}{\Alph{section}}
\section{Appendix}
\label{sec:appendix}

\subsection{Tasks}
\label{sec:tasks}

\begin{table}[h]
\centering
\begin{small}
\begin{tabular}{lcc}
\toprule
Subset & \#Original & \#Filtered \\
\midrule
Algebra & 1,187 & 1,068 \\
Counting \& Probability & 474 & 474 \\
Geometry & 479 & 466 \\
Intermediate Algebra & 903  & 721\\
Number Theory & 540 & 528\\
Prealgebra & 871 & 842\\
Precalculus & 546 & 370 \\
\midrule
\textbf{SUM} & 5,000 & 4,469 \\
\bottomrule
\end{tabular}
\end{small}
\caption{After filtering, the statistics of MATH dataset.}
\label{tab:math}
\end{table}
Appl. comprises the GSM8K~\citep{gsm8k}, SVAMP~\citep{SVAMP}, and Asdiv~\citep{ASDIV} datasets. These datasets contain elementary-level math problems set in specific application scenarios, focusing on mathematical abstraction and modeling skills, with relatively low difficulty. Since they are the same type of questions, we merge them into one task. 
Math, consisting of the transformed MATH~\citep{MATH} dataset, whose answers to the problems are expressed using LaTeX. It's too hard to construct prompts in other languages that meet all the requirements, we select those that can be calculated to a single number, excluding problems with interval or formula-based answers. The filtered results are shown in Table~\ref{tab:math}.

\begin{table}[h]
\centering
\begin{small}
\begin{tabular}{lcc}
\toprule
Task & \#Data & \#Shots\\
\midrule
Appl. & 4,415 & 3 \\
Math	& 4,469 & 3 \\
Date & 369 & 6 \\
Tabular & 149 & 3 \\
Spatial & 2,000 & 3 \\
\bottomrule
\end{tabular}
\end{small}
\caption{Summarization of selected reasoning tasks.}
\label{tab:reasoning_tasks}
\end{table}
Here are the details of our selected tasks, including the number of questions in each task (\#Data) and the number of shots in demonstrations. 

\subsection{Additional Data}
\label{sec:appendix_results}

\begin{table*}[t]
\centering
\begin{small}
\begin{tabular}{cl|cccccc}
\toprule
&  & Appl. & Math & Date & Tabular & Spatial & \textbf{AVG} \\
\midrule
\multirow{5}{*}{\textbf{Starcoder}}
    & Python & 43.06 & 15.78 & 32.79 & 74.50 & 63.55 & 45.94 \\
    & R & 40.63 & 14.63 & \textbf{34.96} & 77.85 & 52.60 & 44.13 \\
    & C++ & 44.21 & 14.43 & 18.43 & 77.18 & 61.90 & 43.23 \\
    & Java & 43.87 & 14.39 & 31.98 & \textbf{81.21} & 60.40 & 46.37  \\
    & JavaScript & \textbf{45.64} & \textbf{17.30} & 32.79 & 74.50 & \textbf{63.65} & \textbf{46.78} \\
\midrule
\multirow{5}{*}{\textbf{Code Llama}}
    & Python & 65.14 & 23.09 & 51.76 & 89.26 & 74.60 & 60.77 \\
    & R  & 63.44 & 23.58 & \textbf{57.99} & 89.93 & 71.35 & 61.26 \\
    & C++  & \textbf{68.79} & 22.76 & 39.57 & 88.59 & 74.90 & 58.92 \\
    & Java  & 67.38 & \textbf{24.84} & 55.28 & \textbf{91.28} & \textbf{82.55} & \textbf{64.27} \\
    & JavaScript  & 65.84 & 23.45 & 46.07 & 85.91 & 76.90 & 59.63 \\
\midrule
\multirow{5}{*}{\textbf{Deepseek Coder}}
    & Python  & 67.34 & 32.00 & 42.55 & 85.23 & 84.45 & 62.31 \\
    & R  & 67.04 & 29.60 & \textbf{50.14} & 88.59 & 89.65 & \textbf{65.00} \\
    & C++ & \textbf{69.40} & 30.63 & 40.38 & \textbf{93.29} & \textbf{90.80} & 64.90 \\
    & Java & 69.08 & 32.02 & 44.17 & 91.28 & 84.50 & 64.21 \\
    & JavaScript  & 68.95 & \textbf{32.29} & 49.59 & 91.28 & 74.20 & 63.26 \\
\bottomrule
\end{tabular}
\end{small}
\caption{The greedy decoding results of each PL of each Code LLMs. The detailed numerical data for Figure \ref{fig:greedy_results}. }
\label{tab:re1}
\end{table*}

Table~\ref{tab:re1} is the raw data of Figure~\ref{fig:greedy_results}, shows the greedy decoding results of each PL of each Code LLMs. Table~\ref{tab:languages_avg_sc_results} shows that on the average performance of three Code LLMs, MultiPoT surpasses all Self-Consistency on four tasks, and is only lower slightly than C++ on Spatial.

\begin{table*}[t]
\centering
\begin{small}
\begin{tabular}{lccccc}
\toprule
& Appl. & Math & Date & Table & Spatial \\
\midrule
Python     & 62.11 & 28.43 & 43.45 & 88.59 & 79.12 \\
R          & 60.08 & 25.99 & 49.50 & 88.14 & 79.18 \\
C++        & 63.66 & 25.23 & 33.88 & 90.38 & \textbf{81.60} \\
Java       & 63.39 & 26.68 & 49.23 & 88.81 & 80.02 \\
JavaScript & 63.09 & 26.97 & 46.43 & 87.02 & 79.80 \\
\midrule
MultiPoT   & \textbf{64.39} & \textbf{28.64} & \textbf{51.13} & \textbf{92.17} & 80.95 \\
\bottomrule
\end{tabular}
\end{small}
\caption{The average performance of three Code LLMs for Self-Consistency and MultiPoT in each task.}
\label{tab:languages_avg_sc_results}
\end{table*}

\begin{table*}[t]
\centering
\begin{small}
\resizebox{\textwidth}{!}{
\begin{tabular}{l|ccc|ccc|ccc|ccc|ccc}
\toprule
& \multicolumn{3}{c|}{Appl.} & \multicolumn{3}{c|}{Math} & \multicolumn{3}{c|}{Date} 
& \multicolumn{3}{c|}{Tabular} & \multicolumn{3}{c}{Spatial}\\
\cmidrule{2-16}
& AC & WA & RE & AC & WA & RE & AC & WA & RE & AC & WA & RE & AC & WA & RE\\

\midrule
    Python & 67.34 & 30.06 & 2.60 & 32.00 & 48.00 & \multicolumn{1}{r|}{20.00} & 42.55 & 57.18 & 0.27 
    &85.23 & 6.04 & 8.72 & 84.45 & \multicolumn{1}{r}{12.65} & 2.90\\
    R & 67.04 & 31.51 & 1.45 & 29.60 & 53.79 & \multicolumn{1}{r|}{16.60} & 50.14 & 43.63 & 6.23
    & 88.59 & 8.05 & 3.36 & 89.65 & \multicolumn{1}{r}{6.65} & 3.70\\
    C++ & 69.40 & 30.15 & 0.45 & 30.63 & 61.74 & \multicolumn{1}{r|}{7.63} & 40.38 & 59.62 & 0.00
    & 93.29 & 6.71 & 0.00 & 90.80 & \multicolumn{1}{r}{8.95} & 0.25\\
    Java & 69.08 & 30.28 & 0.63 & 32.02 & 60.80 & \multicolumn{1}{r|}{7.18} & 44.17 & 47.15 & 8.67
    & 91.28 & 7.38 & 1.34 & 84.50 & \multicolumn{1}{r}{14.65} & 0.85\\
    JavaScript & 68.95 & 28.99 & 2.06 & 32.29 & 55.36 & \multicolumn{1}{r|}{12.35} & 49.59 & 49.86 & 0.54
    & 91.28 & 7.38 & 1.34 & 74.20 & \multicolumn{1}{r}{19.45} & 6.35\\
\bottomrule
\end{tabular}
}
\end{small}
\caption{The execution result of programs generated from Deepseek Coder for five languages on five tasks. \textbf{AC} represents \textbf{Accept}, which means the program can generate a correct answer. \textbf{Wrong} means the answer is not right. \textbf{RE} represents \textbf{Runtime Error}, which means the program does not execute normally.}
\label{tab:dc_error_analyze}
\end{table*}

\begin{table*}[t]
\centering
\begin{small}
\resizebox{\textwidth}{!}{
\begin{tabular}{l|ccc|ccc|ccc|ccc|ccc}
\toprule
& \multicolumn{3}{c|}{Appl.} & \multicolumn{3}{c|}{Math} & \multicolumn{3}{c|}{Date} 
& \multicolumn{3}{c|}{Tabular} & \multicolumn{3}{c}{Spatial}\\
\cmidrule{2-16}
& AC & WA & RE & AC & WA & RE & AC & WA & RE & AC & WA & RE & AC & WA & RE\\

\midrule
    Python & 43.06 & 53.70 & 3.24 & 15.78 & 60.66 & 23.56 & 32.79 & 63.41 & 3.79 
    & 74.50 & 14.09 & \multicolumn{1}{r|}{11.41} & 63.55 & 29.65 & \multicolumn{1}{r}{6.80}\\
    R & 40.63 & 57.94 & 1.43 & 14.63 & 66.08 & 19.29 & 34.96 & 55.83 & 9.21
    & 77.85 & 19.46 & \multicolumn{1}{r|}{2.68} & 52.60 & 28.65 & \multicolumn{1}{r}{18.75}\\
    C++ & 44.21 & 54.74 & 1.04 & 14.43 & 71.81 & 13.76 & 18.43 & 81.57 & 0.00
    & 77.18 & 18.12 & \multicolumn{1}{r|}{4.70} & 61.90 & 37.75 & \multicolumn{1}{r}{0.35}\\
    Java & 43.87 & 54.65 & 1.47 & 14.39 & 74.71 & 10.90 & 31.98 & 61.52 & 6.50
    & 81.21 & 17.45 & \multicolumn{1}{r|}{1.34} & 60.40 & 31.80 & \multicolumn{1}{r}{7.80}\\
    JavaScript & 45.64 & 52.21 & 2.15 & 17.30 & 66.68 & 16.02 & 32.79 & 67.21 & 0.00 & 74.50 & 24.16 & \multicolumn{1}{r|}{1.34} & 63.65 & 30.10 & \multicolumn{1}{r}{6.25}\\
\bottomrule
\end{tabular}
}
\end{small}
\caption{The execution result of programs generated from Starcoder.}
\label{tab:sc_error_analyze}
\end{table*}

\begin{table*}[t]
\centering
\begin{small}
\resizebox{\textwidth}{!}{
\begin{tabular}{l|ccc|ccc|ccc|ccc|ccc}
\toprule
& \multicolumn{3}{c|}{Appl.} & \multicolumn{3}{c|}{Math} & \multicolumn{3}{c|}{Date} 
& \multicolumn{3}{c|}{Tabular} & \multicolumn{3}{c}{Spatial}\\
\cmidrule{2-16}
& AC & WA & RE & AC & WA & RE & AC & WA & RE & AC & WA & RE & AC & WA & RE\\

\midrule
    Python & 65.14 & 32.14 & 2.72 & 23.09 & 57.04 & \multicolumn{1}{r|}{19.87} & 51.76 & 48.24 & 0.00 & 89.26 & \multicolumn{1}{r}{8.72} & 2.01 & 73.60 & 18.85 & 7.55\\
    R & 63.44 & 34.47 & 2.08 & 23.58 & 61.42 & \multicolumn{1}{r|}{14.99} & 57.99 & 41.73 & 0.27 & 89.93 & \multicolumn{1}{r}{9.40} & 0.67 & 71.35 & 24.00 & 4.65\\
    C++ & 68.79 & 30.87 & 0.34 & 22.76 & 71.69 & \multicolumn{1}{r|}{5.55} & 39.57 & 59.89 & 0.54 & 88.59 & \multicolumn{1}{r}{10.74} & 0.67 & 74.90 & 23.80 & 1.30\\
    Java & 67.38 & 32.07 & 0.54 & 24.84 & 68.20 & \multicolumn{1}{r|}{6.96} & 55.28 & 38.75 & 5.96 & 91.28 & \multicolumn{1}{r}{6.71} & 2.01 & 82.55 & 17.05 & 0.40\\
    JavaScript & 65.84 & 32.41 & 1.74 & 23.45 & 67.69 & \multicolumn{1}{r|}{8.86} & 46.07 & 53.39 & 0.54 & 85.91 & \multicolumn{1}{r}{12.75} & 1.34 & 76.90 & 21.50 & 1.60\\
\bottomrule
\end{tabular}
}
\end{small}
\caption{The execution result of programs generated from Code Llama.}
\label{tab:cl_error_analyze}
\end{table*}

\begin{table*}[t]
\centering
\begin{small}
\resizebox{\textwidth}{!}{
\begin{tabular}{l|ccc|ccc|ccc|ccc|ccc}
\toprule
& \multicolumn{3}{c|}{Appl.} & \multicolumn{3}{c|}{Math} & \multicolumn{3}{c|}{Date} 
& \multicolumn{3}{c|}{Tabular} & \multicolumn{3}{c}{Spatial}\\
\cmidrule{2-16}
& AC & WA & RE & AC & WA & RE & AC & WA & RE & AC & WA & RE & AC & WA & RE\\

\midrule
    Python & 80.75 &15.61 &3.65 &39.74 &22.76 &37.50 &46.61 &52.85 &0.54 &94.63 &4.70 &0.67 &91.70 &\multicolumn{1}{r}{8.00} &0.30\\
    R &79.37 &16.78 &3.85 &34.86 &25.53 &39.61 &55.01 &42.82 &2.17 &89.93 &7.38 &2.68 &92.85 &\multicolumn{1}{r}{5.75} &1.40\\
    C++ &79.46 &16.67 &3.87 &39.90 &39.94 &20.16 &47.70 &50.95 &1.36 &91.95 &4.03 &4.03 &86.65 &\multicolumn{1}{r}{12.20} &1.15\\
    Java &80.63 &16.44 &2.92 &42.65 &41.96 &15.39 &51.22 &40.92 &7.86 &87.92 &6.71 &5.37 &86.30 &\multicolumn{1}{r}{11.00} &2.70\\
    JavaScript &81.25 &15.24 &3.51 &36.07 &24.23 &39.70 &55.01 &44.17 &0.81 &92.62 &4.70 &2.68 &90.15 &\multicolumn{1}{r}{9.70} &0.15\\
\bottomrule
\end{tabular}
}
\end{small}
\caption{The execution result of programs generated from ChatGPT.}
\label{tab:gpt_error_analyze}
\end{table*}

\subsection{Error Analysis}
\label{sec:error_analyse}

We further classify incorrect results into Wrong Answer (WA) and Runtime Error (RE), representing cases where the program runs but produces incorrect answers and where the program encounters errors during execution, respectively. Tables~\ref{tab:dc_error_analyze} to Table~\ref{tab:gpt_error_analyze} show the results for the four models.

It is evident that there are significant differences in the proportion of runtime errors (RE) across different languages and models for each task. Even languages with similar performance exhibit different distributions of errors. For instance, on Appl. of Deepseek Coder, the accuracy difference between Java and JavaScript is less than 0.1\%, yet JavaScript has an RE rate of 2.06\%, while Java’s is only 0.63\%. It indicates that the types of errors vary significantly among languages.

A further categorization of the types of RE is conducted. We classify all REs into eight error types. \textbf{Redeclaration} represents duplicate naming of variables. \textbf{Division by Zero} represents the denominator in the division is zero. \textbf{Illegal Output} represents the answer can not be parsed or converted correctly. \textbf{Time Limit Error} represents the program runs out of time and sometimes it is due to stack space overflow. \textbf{Compile Error} often means there are some syntax error in the program. \textbf{Undefined Identifier} includes Undefined Variables and Undefined Functions, which means the variables or functions are not defined before they are used. \textbf{Variable Type Error} indicates that the types of variables are mismatched when they are involved in some operations, for example addition or division. Table~\ref{tab:dc_re} shows the proportion of different RE types for Deepseek Coder across five tasks and five languages. Table~\ref{tab:appl_re} presents the proportion of various RE types for four LLMs on Appl. across all languages. Deepseek Coder and the Appl. task are selected because the languages have the most similar performance on them. The results demonstrate that even in scenarios where languages exhibit similar performance, the proportions of RE differ significantly among languages. For instance, the RE rate on ChatGPT's Appl. of R and C++ differs by only 0.02\%, yet Illegal Output account for 82.46\% of C++ errors, in comparison to only 24.71\% for R. Given that each prompt is accurate, the differing error distributions are attributable to the intrinsic characteristics of the languages, thereby demonstrating their diversity and the non-repetitive nature of their errors.

Upon further analysis of generated contents, several common failure patterns emerge:
\begin{itemize} 
    \item \textbf{Date Calculation}: PoTs often misinterpret the difference between two dates as exclusive, contrary to natural language conventions, where the interval is typically inclusive. Nevertheless, R demonstrates comparable performance to CoT on Date, indicating its potential for temporal reasoning tasks.

    \item \textbf{Output Content}: PoTs frequently respond to yes/no questions by outputting attributes instead of directly answering the question. For example, when asked, `On the desk, there is a teal pen and a yellow textbook. Is the textbook yellow?', the correct answer is `yes', but PoT might respond with `yellow'.

    \item \textbf{Demonstration Constraint}: Demonstrations are less restrictive for JavaScript, as it tends to output extra information beyond what is required, including descriptive sentences and variables, even when only the final answer is needed.

    \item \textbf{Syntax Preference}: C++ and Java tend to leverage language-specific constructs like for-loops, often reaching correct solutions in pattern recognition problems. In contrast, other languages that attempt step-by-step calculations may add or omit steps, leading to errors.

    \item \textbf{Resource Constraints}: Low-resource languages like R may cause the model to call non-existent packages, particularly in complex mathematical problems that rely on external tools.
\end{itemize}


\begin{table*}[t]
\centering
\begin{small}
\resizebox{\textwidth}{!}{
\begin{tabular}{cl|rrrrrrrr}
\toprule
Task & Language & \makecell[c]{Redecl\\aration} & \makecell[c]{Division\\by Zero} 
& \makecell[c]{Illegal\\Output} 
& \makecell[c]{Time Limit\\Error} & \makecell[c]{Compile\\Error} 
& \makecell[c]{Undefined\\Identifier} 
& \makecell[c]{Variable\\Type Error} & \makecell[c]{Other\\Error} \\
\midrule
\multirow{5}{*}{\textbf{Appl.}}
    & Python & - & - & 61.74 & 2.61 & 9.57 & 23.48 & 2.61 & - \\
    & R & - & - & 32.81 & 4.69 & 39.06 & 23.44 & - & -\\
    & C++ & 60.00 & - & 15.00 & 10.00 & 5.00 & 5.00 & - & 5.00\\
    & Java & 46.43 & 7.14 & - & 3.57 & 14.29 & 3.57 & 25.00 & - \\
    & JavaScript & 5.49 & - & 84.62 & 2.20 & 2.20 & 5.49 & - & -\\
\midrule
\multirow{5}{*}{\textbf{Math}}
    & Python & - & 2.57 & 19.91 & 31.21 & 4.70 & 31.1 & 7.61 & 2.91\\
    & R  & - & - & 20.22 & 10.65 & 10.24 & 38.27 & 1.35 & 19.27\\
    & C++  & 2.93 & - & 17.60 & 28.15 & 21.11 & 7.04 & 4.40 & 18.77\\
    & Java  & 1.25 & 3.12 & 16.20 & 28.97 & 16.51 & 8.72 & 3.12 & 22.12\\
    & JavaScript  & 1.09 & - & 23.01 & 22.64 & 6.34 & 32.43 & - & 14.49\\
\midrule
\multirow{5}{*}{\textbf{Date}}
    & Python  & - & - & - & - & - & 100 & - & -\\
    & R  & - & - & - & 95.65 & 4.35 & - & - & -\\
    & C++ & - & - & - & - & - & - & - & - \\
    & Java & - & - & - & - & 3.12 & 56.25 & - & 40.62 \\
    & JavaScript  & 50.00 & - & - & - & -  & 50.00 & - & - \\
\midrule
\multirow{5}{*}{\textbf{Tabular}}
    & Python & - & - & - & - & 84.62 & - & 7.69 & 7.69 \\
    & R & - & - & - & - & - & - & 20.00 & 80.00 \\
    & C++ & - & - & - & - & - & - & - & - \\
    & Java & - & - & - & - & 50.00 & - & - & 50.00 \\
    & JavaScript & - & - & - & - & - & 100 & - & - \\
\midrule
\multirow{5}{*}{\textbf{Spatial}}
    & Python & - & - & - & - & 1.72 & 1.72 & 96.55 & - \\
    & R & - & - & - & - & - & - & 22.97 & 77.03 \\
    & C++ & - & - & 20.00 & - & 80.00 & - & - & - \\
    & Java & - & - & - & - & 5.88 & - & 17.65 & 76.47 \\
    & JavaScript & - & - & - & - & 0.79 & 96.85 & - & 2.36 \\

\bottomrule
\end{tabular}
}
\end{small}
\caption{Runtime Error concrete analysis for five languages on five tasks of Deepseek Coder.}
\label{tab:dc_re}
\end{table*}

\begin{table*}[t]
\centering
\begin{small}
\resizebox{\textwidth}{!}{
\begin{tabular}{cl|rrrrrrrr}
\toprule
Model & Language & \makecell[c]{Redecl\\aration} & \makecell[c]{Division\\by Zero} 
& \makecell[c]{Illegal\\Output} 
& \makecell[c]{Time Limit\\Error} & \makecell[c]{Compile\\Error} 
& \makecell[c]{Undefined\\Identifier} 
& \makecell[c]{Variable\\Type Error} & \makecell[c]{Other\\Error} \\
\midrule
\multirow{5}{*}{\textbf{Starcoder}}
    & Python & - & - & 69.93 & 1.40 & 3.50 & 17.48 & 1.40 & - \\
    & R & - & - & 38.10 & 1.59 & 23.81 & 34.92 & 1.59 & -\\
    & C++ & 28.26 & - & 8.70 & 17.39 & 17.39 & 8.70 & - & 19.57\\
    & Java & 6.15 & 3.08 & 3.08 & 3.08 & 24.62 & 3.08 & 56.92 & - \\
    & JavaScript & 29.47 & - & 53.68 & 3.16 & 2.11 & 9.47 & - & 2.11\\
\midrule
\multirow{5}{*}{\textbf{Code Llama}}
    & Python & - & - & 49.17 & 1.67 & 6.67 & 39.17 & 1.67 & 1.67\\
    & R  & - & - & 36.96 & 3.26 & 4.35 & 51.09 & 1.09 & 3.26\\
    & C++  & 13.33 & - & 6.67 & 6.67 & 20.00 & 33.33 & - & 20.00\\
    & Java  & 8.33 & - & 4.17 & - & 12.50 & 8.33 & 58.33 & 8.33\\
    & JavaScript  & 9.09 & - & 68.83 & 1.30 & 2.60 & 14.29 & - & 3.90\\
\midrule
\multirow{5}{*}{\textbf{Deepseek Coder}}
    & Python & - & - & 61.74 & 2.61 & 9.57 & 23.48 & 2.61 & - \\
    & R & - & - & 32.81 & 4.69 & 39.06 & 23.44 & - & -\\
    & C++ & 60.00 & - & 15.00 & 10.00 & 5.00 & 5.00 & - & 5.00\\
    & Java & 46.43 & 7.14 & - & 3.57 & 14.29 & 3.57 & 25.00 & - \\
    & JavaScript & 5.49 & - & 84.62 & 2.20 & 2.20 & 5.49 & - & -\\
\midrule
\multirow{5}{*}{\textbf{ChatGPT}}
    & Python & - & - & 51.55 & 0.62 & 10.56 & 35.40 & 1.24 & 0.62 \\
    & R & - & - & 24.71 & 0.59 & 23.53 & 43.53 & 1.18 & 1.76 \\
    & C++ & - & 0.58 & 82.46 & 4.68 & - & 8.19 & - & 4.09 \\
    & Java & 0.78 & 2.33 & 67.44 & 2.33 & 1.55 & 9.3 & 13.95 & 2.33 \\
    & JavaScript & 2.58 & - & 50.32 & 1.29 & 0.65 & 36.13 & - & 9.03 \\

\bottomrule
\end{tabular}
}
\end{small}
\caption{Runtime Error concrete analysis for five languages on Appl. of four LLMs.}
\label{tab:appl_re}
\end{table*}


\subsection{Difference Between Code Generation and PoT}
\label{sec:task_difference}
Figure~\ref{fig:factors} illustrates that performance in code generation does not fully align with that in reasoning tasks. 

Although both tasks involve generating code to solve problems, their objectives differ. The code generation task assesses the LLM's ability to assist development in an engineering environment, covering real-world engineering issues. For example, consider the following problems: 'Given a positive floating point number, return its decimal part' and 'Given a list of integers, return a tuple containing the sum and product of all the integers in the list.' Although these problems require some reasoning, the focus is primarily on language comprehension and engineering skills.

In contrast, reasoning tasks aim to test the LLM's logical reasoning abilities. The generated code acts as a carrier of logic and facilitates the use of tools, such as more precise calculations, dictionaries for storing and retrieving attribute information, or calendars to aid in date reasoning. Reasoning tasks focus on a subset of a programming language's capabilities, rather than its entire spectrum in engineering practice.

Therefore, although there is some overlap between code generation and reasoning tasks, they are not entirely the same. This is why there is only partial consistency between the two tasks in Figure~\ref{fig:factors} and highlights the necessity of testing different programming languages in reasoning tasks..

\clearpage
\subsection{Prompts}
\label{sec:prompt}
Here are our multilingual prompts. We show prompts of Tabular(3-shots) as an example and prompts for other tasks are in the released code.

\begin{figure*}[h]
\centering
\includegraphics[width=\textwidth]{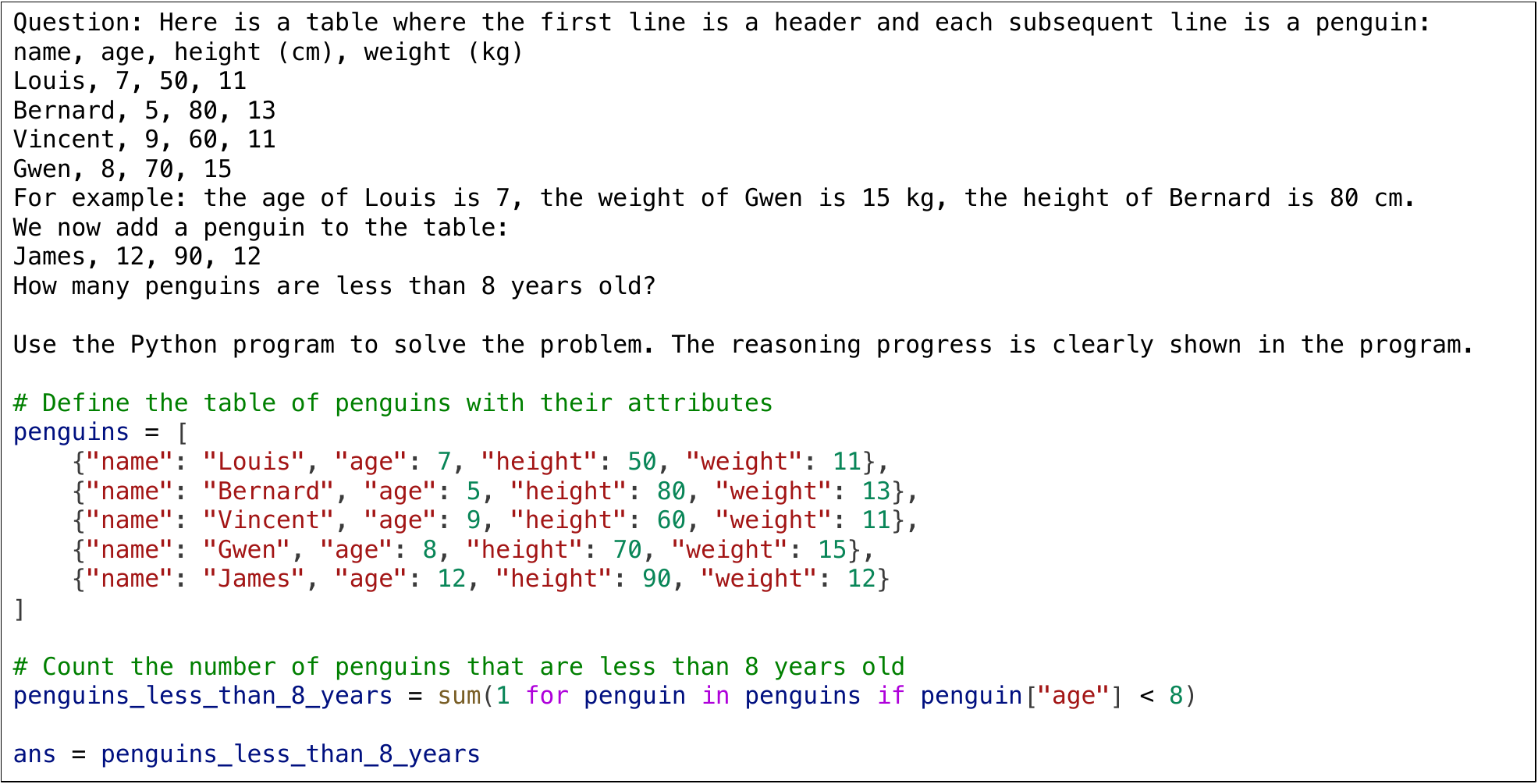}
\caption{Python Prompt of the first question.}
\label{fig:py_q1_prompt}
\end{figure*}

\begin{figure*}[h]
\centering
\includegraphics[width=\textwidth]{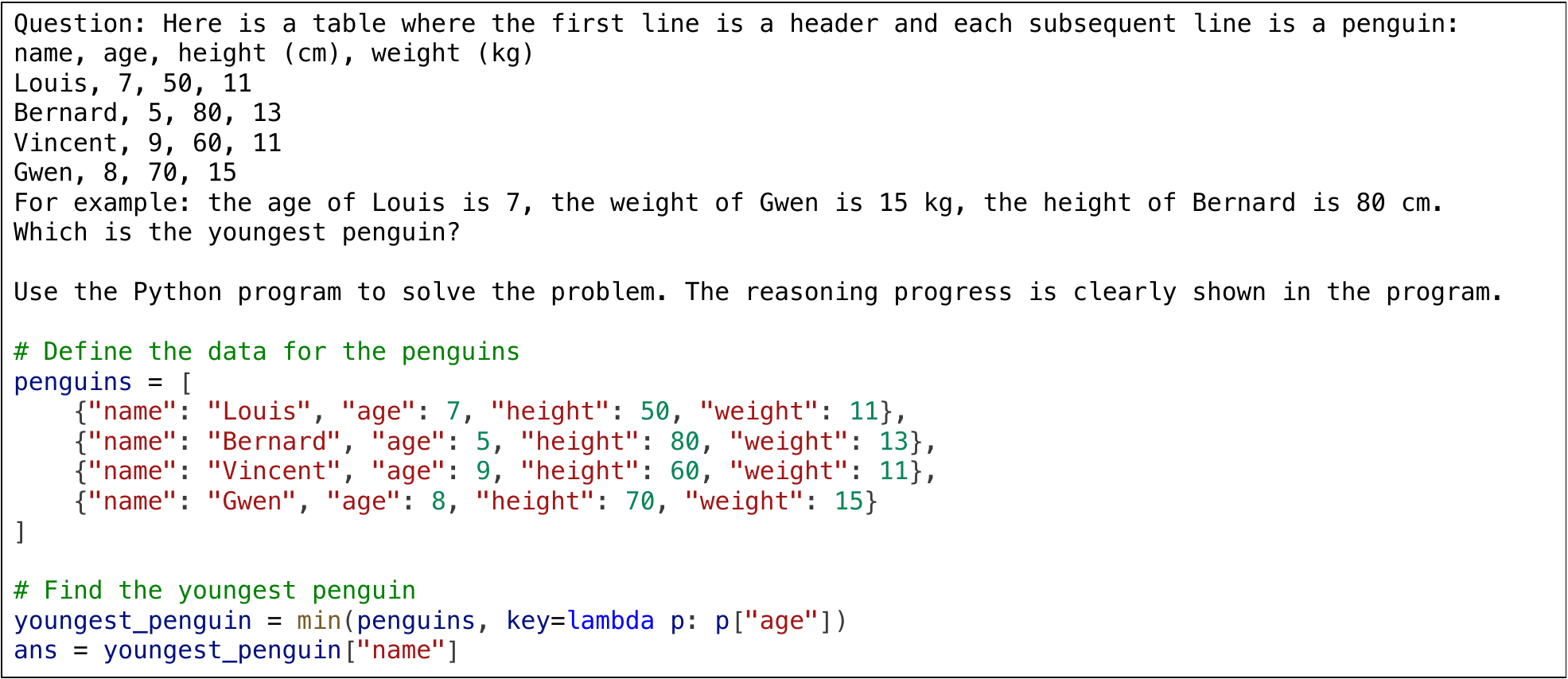}
\caption{Python Prompt of the second question.}
\label{fig:py_q2_prompt}
\end{figure*}

\begin{figure*}[h]
\centering
\includegraphics[width=\textwidth]{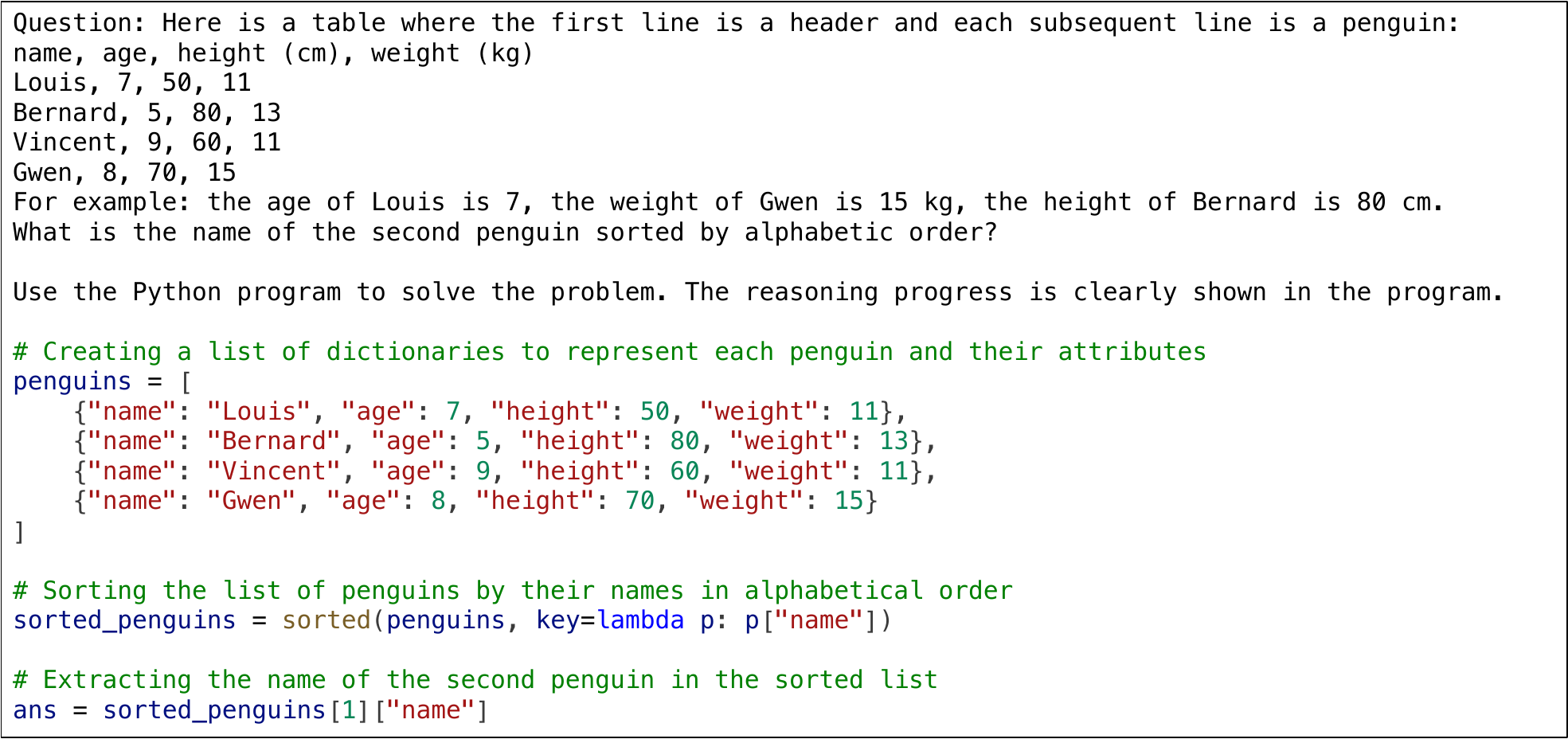}
\caption{Python Prompt of the third question.}
\label{fig:py_q3_prompt}
\end{figure*}

\begin{figure*}[h]
\centering
\includegraphics[width=\textwidth]{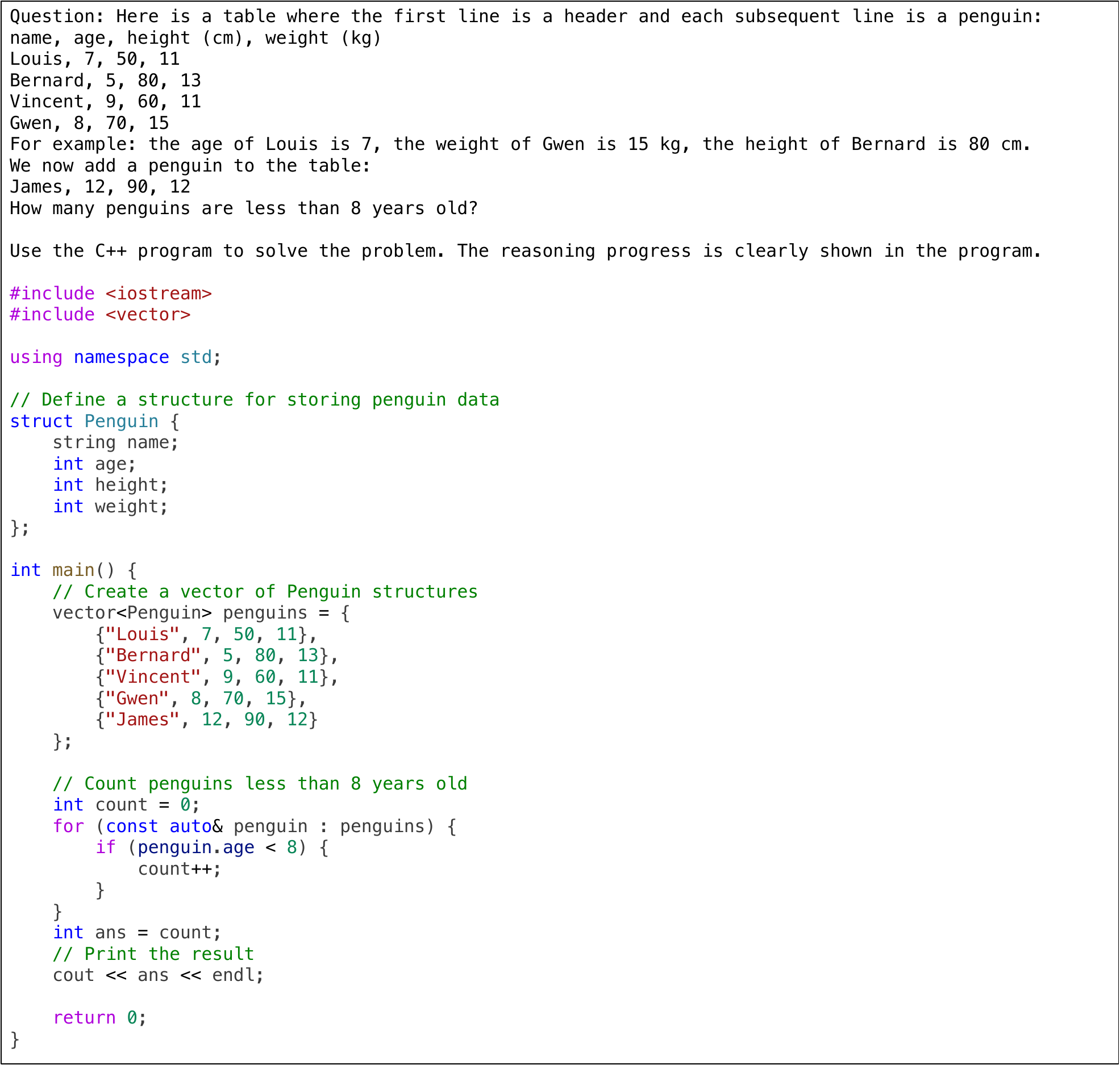}
\caption{C++ Prompt of the first question.}
\label{fig:cpp_q1_prompt}
\end{figure*}

\begin{figure*}[h]
\centering
\includegraphics[width=\textwidth]{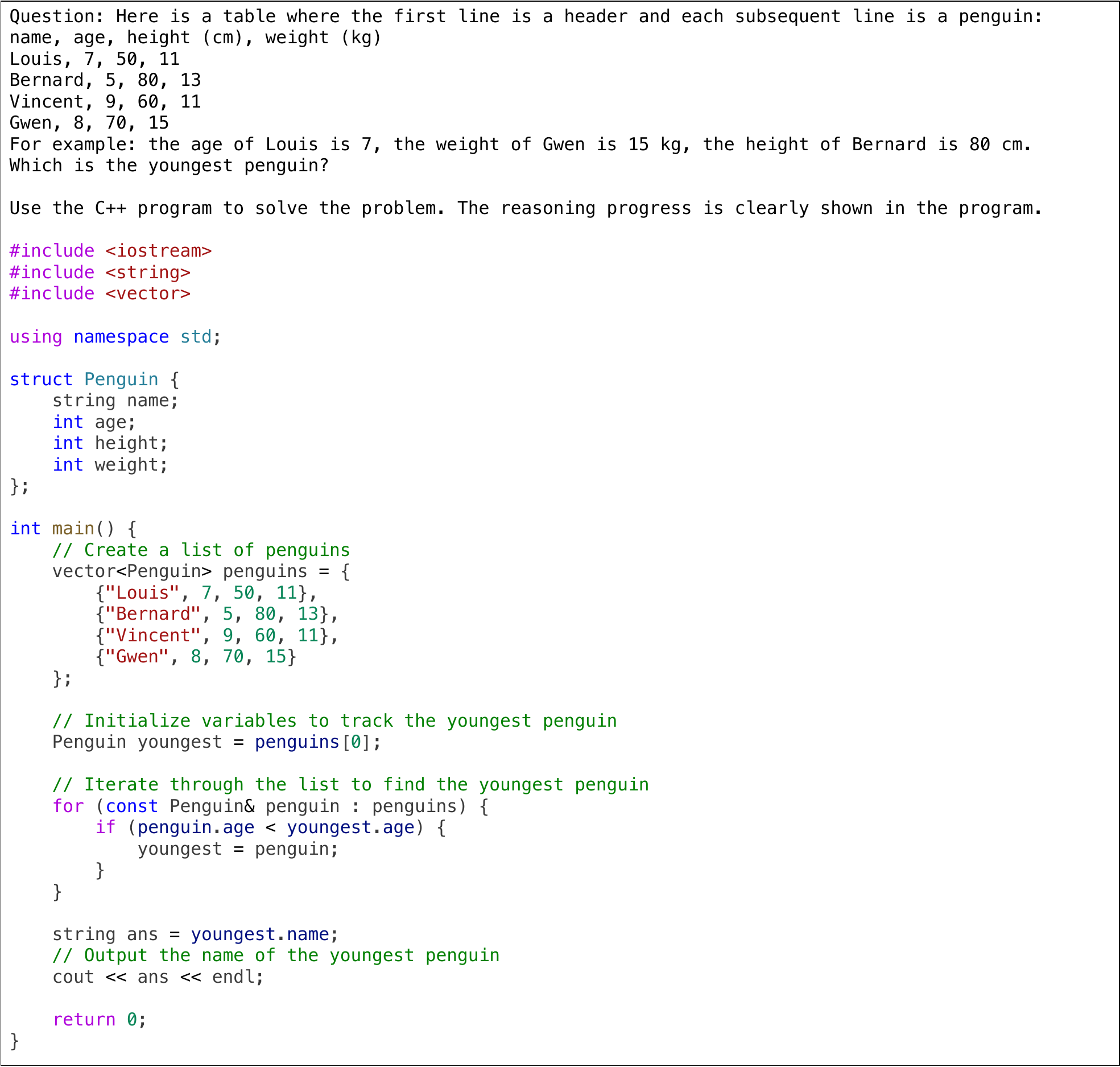}
\caption{C++ Prompt of the second question.}
\label{fig:cpp_q2_prompt}
\end{figure*}

\begin{figure*}[h]
\centering
\includegraphics[width=\textwidth]{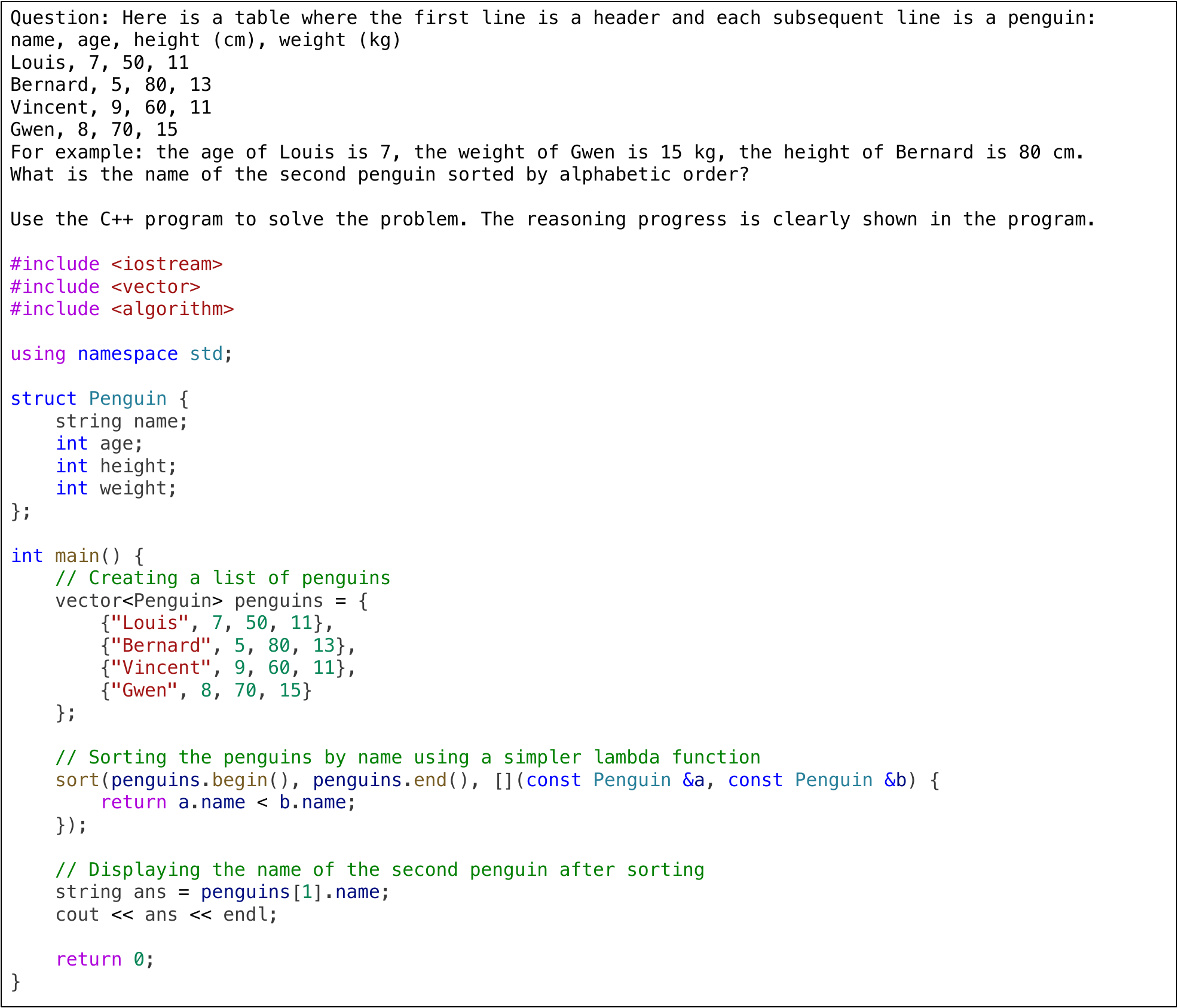}
\caption{C++ Prompt of the third question.}
\label{fig:cpp_q3_prompt}
\end{figure*}

\begin{figure*}[h]
\centering
\includegraphics[width=\textwidth]{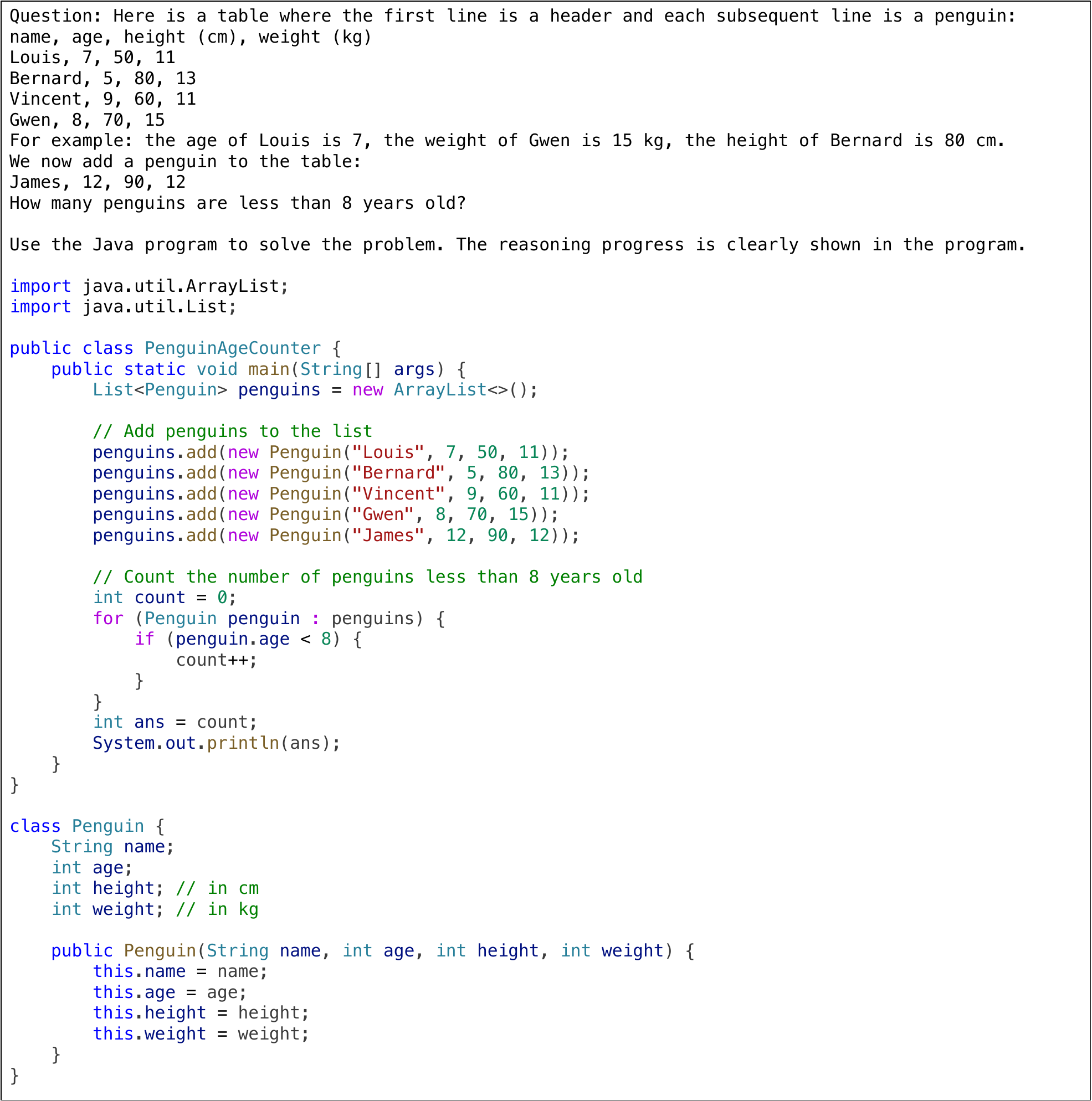}
\caption{Java Prompt of the first question.}
\label{fig:cpp_prompt}
\end{figure*}

\begin{figure*}[h]
\centering
\includegraphics[width=\textwidth]{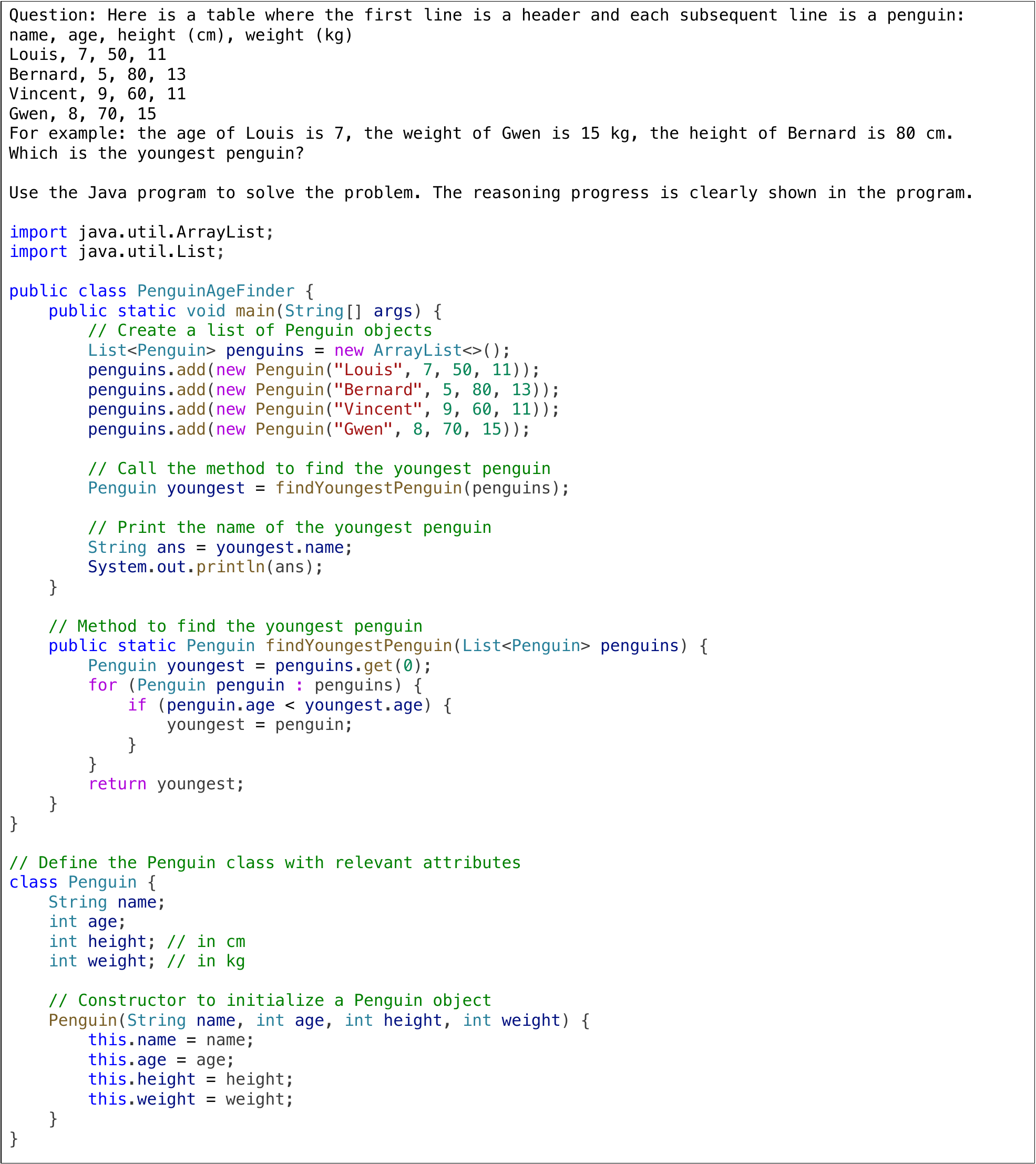}
\caption{Java Prompt of the second question.}
\label{fig:cpp_prompt}
\end{figure*}

\begin{figure*}[h]
\centering
\includegraphics[width=\textwidth]{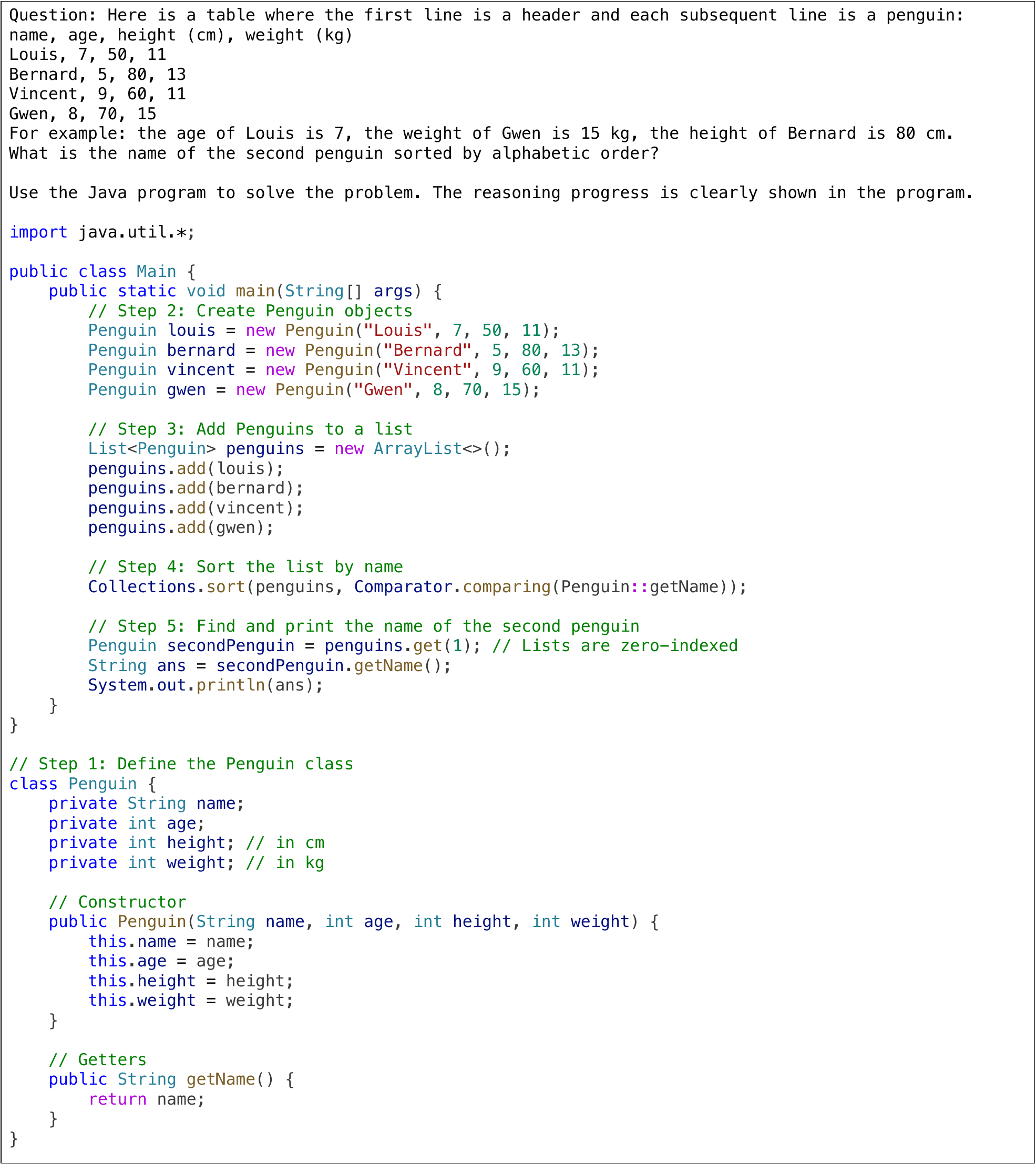}
\caption{Java Prompt of the third question.}
\label{fig:cpp_prompt}
\end{figure*}

\begin{figure*}[h]
\centering
\includegraphics[width=\textwidth]{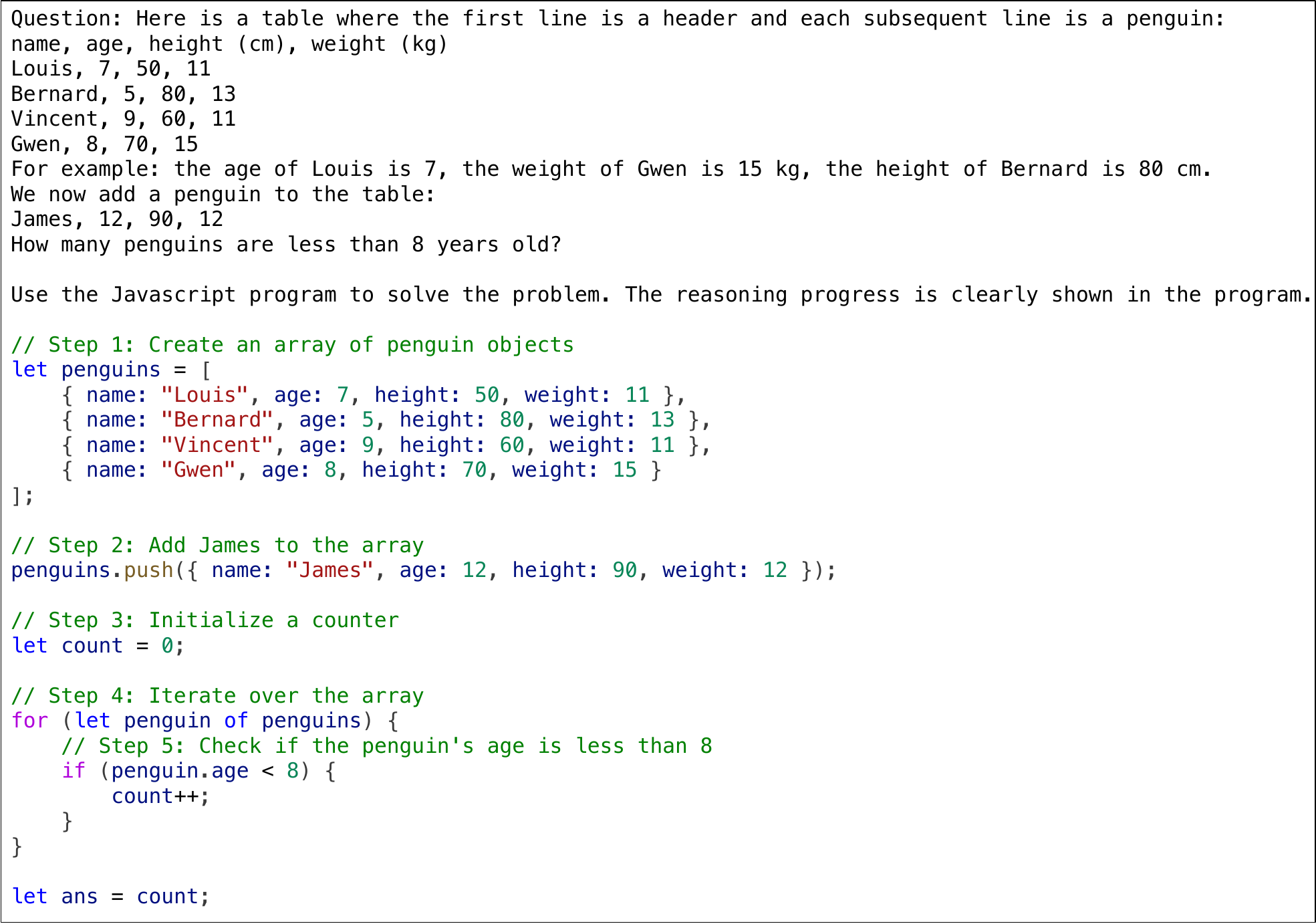}
\caption{JavaScript Prompt of the first question.}
\label{fig:cpp_prompt}
\end{figure*}

\begin{figure*}[h]
\centering
\includegraphics[width=\textwidth]{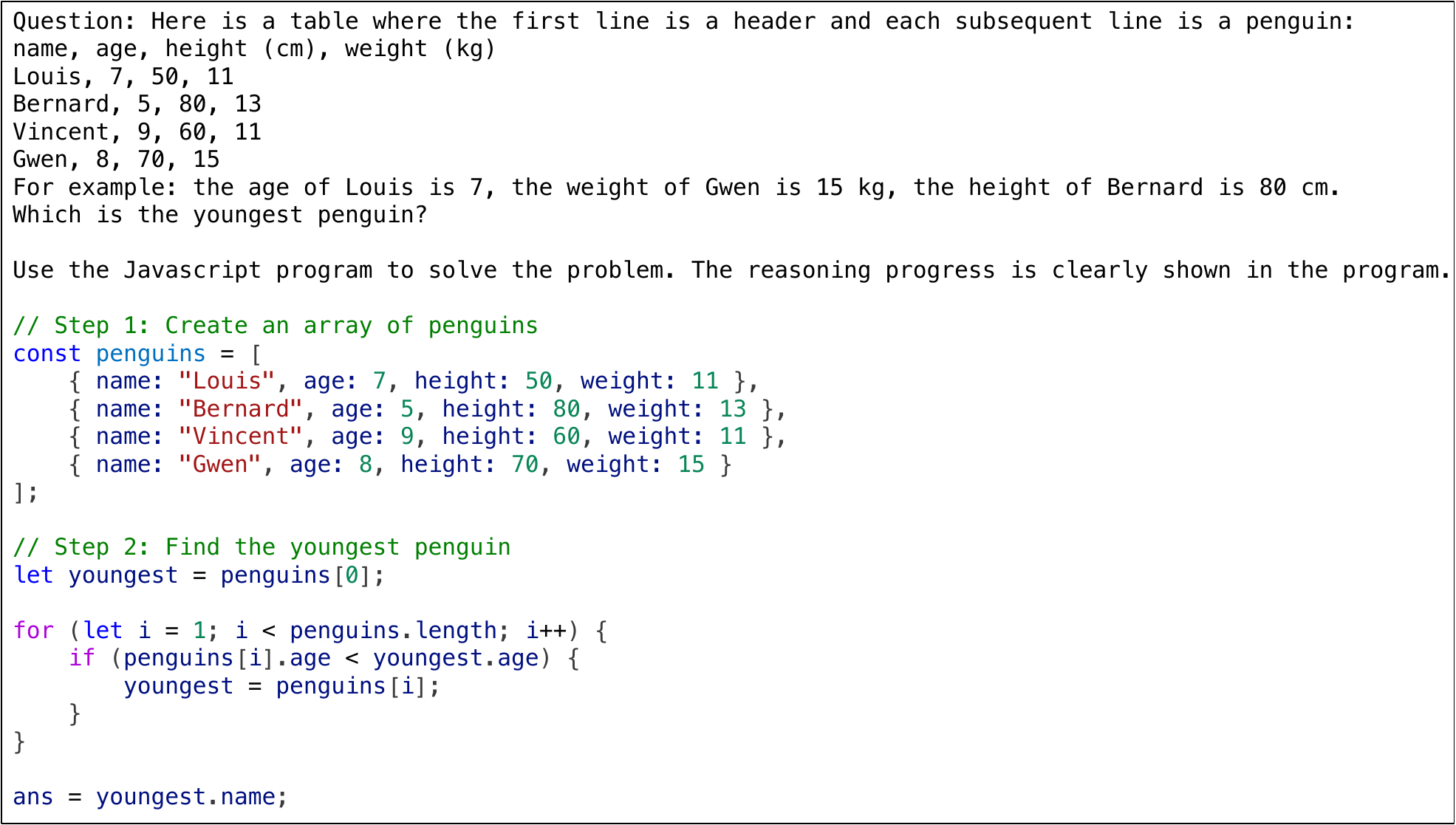}
\caption{JavaScript Prompt of the second question.}
\label{fig:cpp_prompt}
\end{figure*}

\begin{figure*}[h]
\centering
\includegraphics[width=\textwidth]{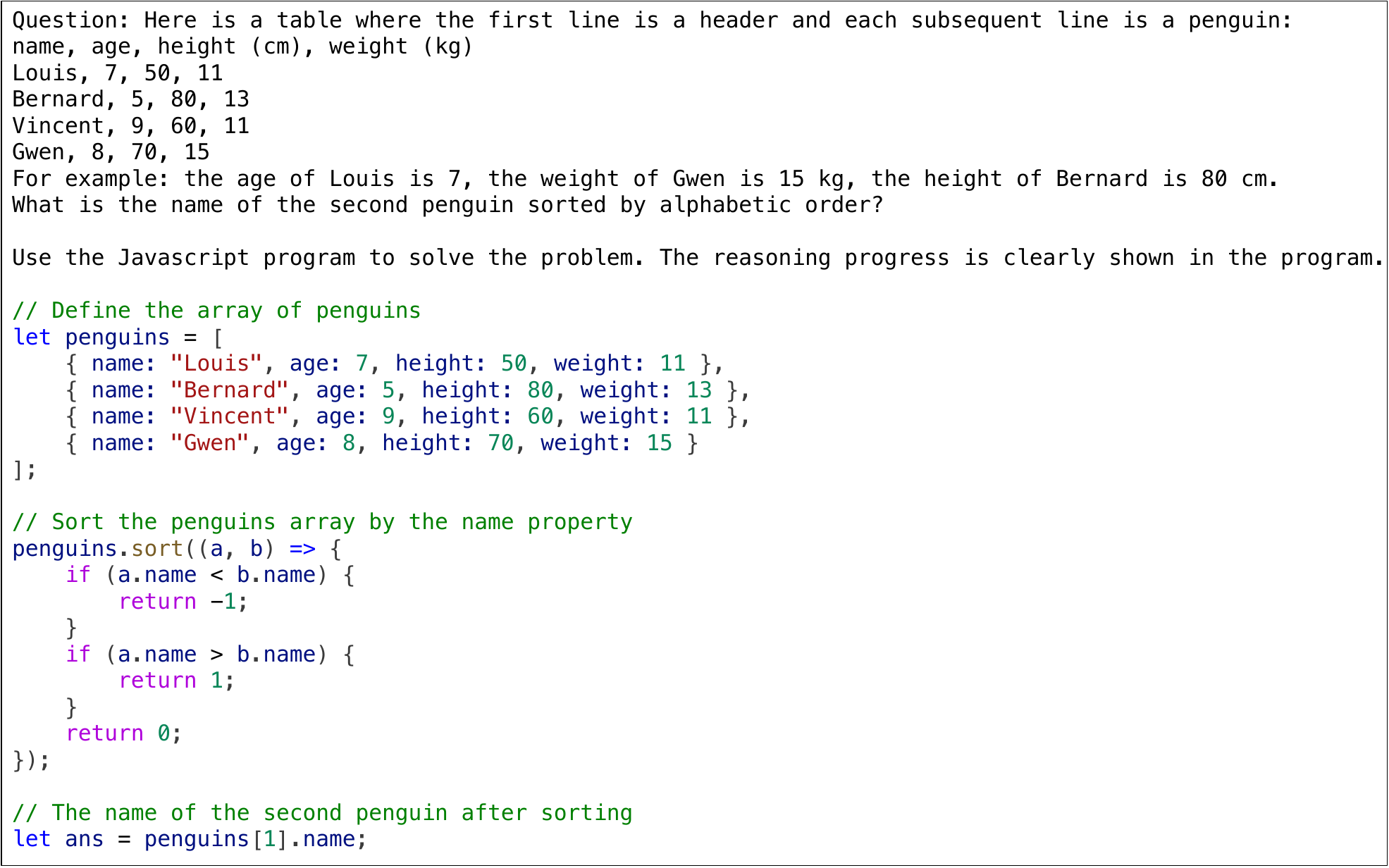}
\caption{JavaScript Prompt of the third question.}
\label{fig:cpp_prompt}
\end{figure*}

\begin{figure*}[h]
\centering
\includegraphics[width=\textwidth]{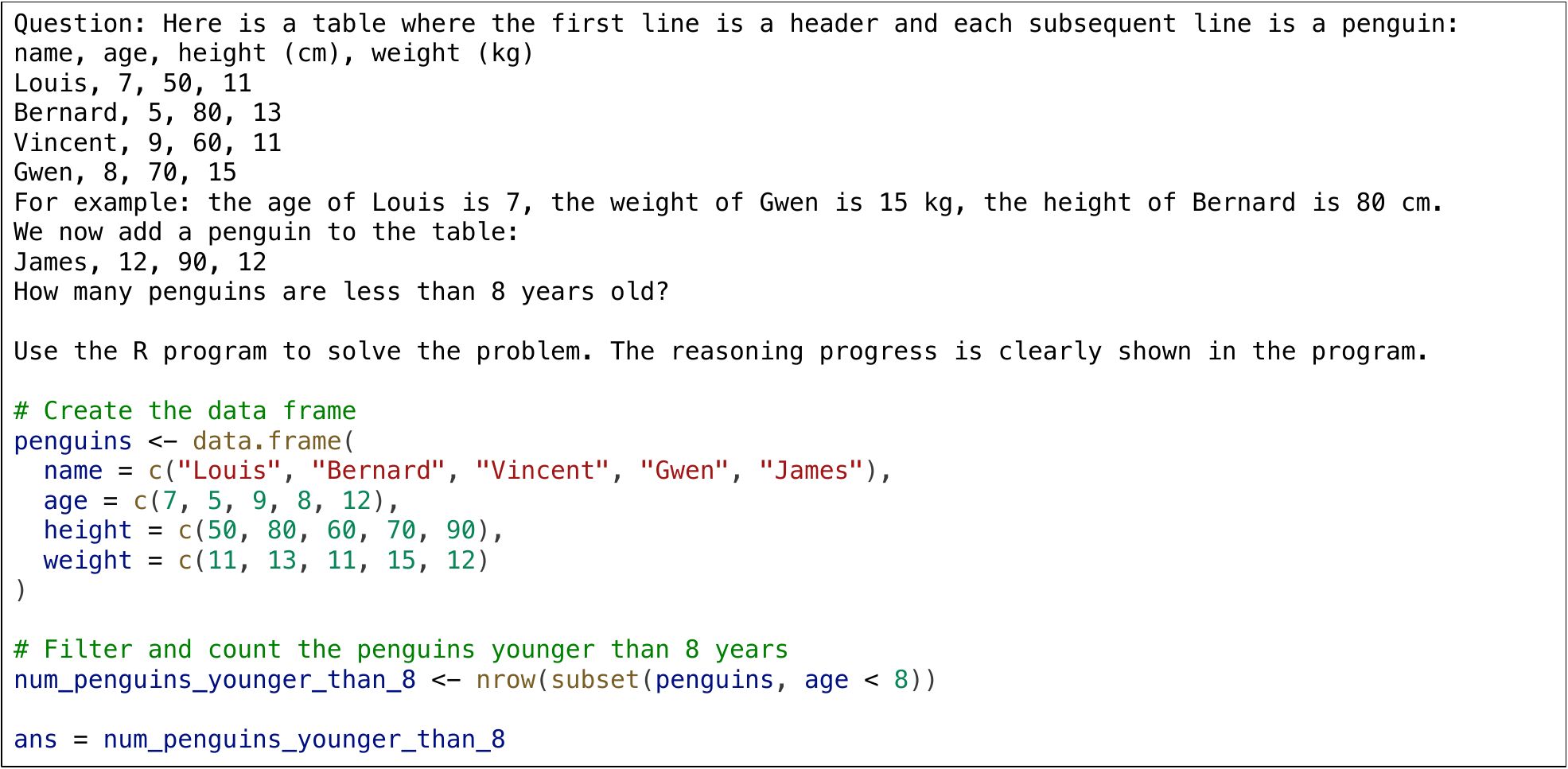}
\caption{R Prompt of the first question.}
\label{fig:cpp_prompt}
\end{figure*}

\begin{figure*}[h]
\centering
\includegraphics[width=\textwidth]{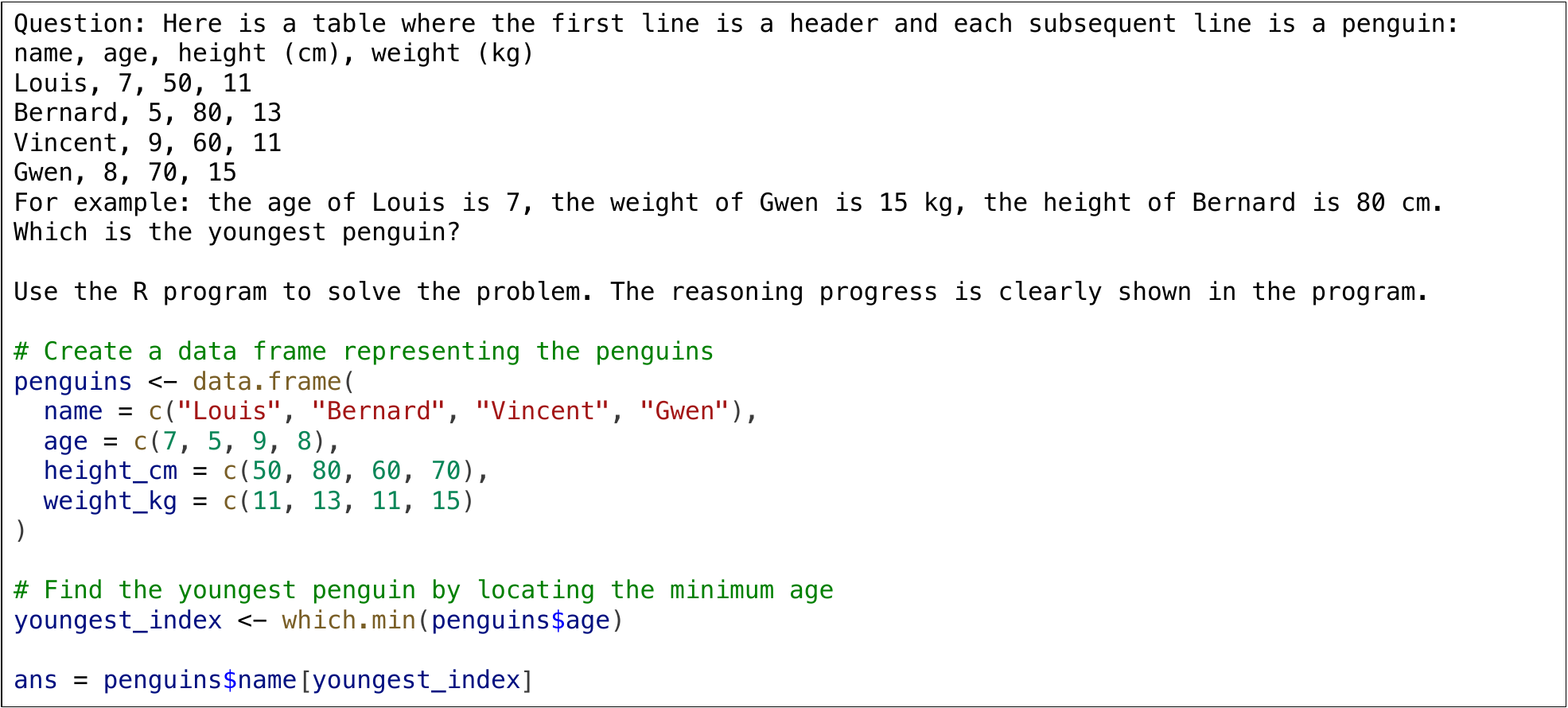}
\caption{R Prompt of the second question.}
\label{fig:cpp_prompt}
\end{figure*}

\begin{figure*}[h]
\centering
\includegraphics[width=\textwidth]{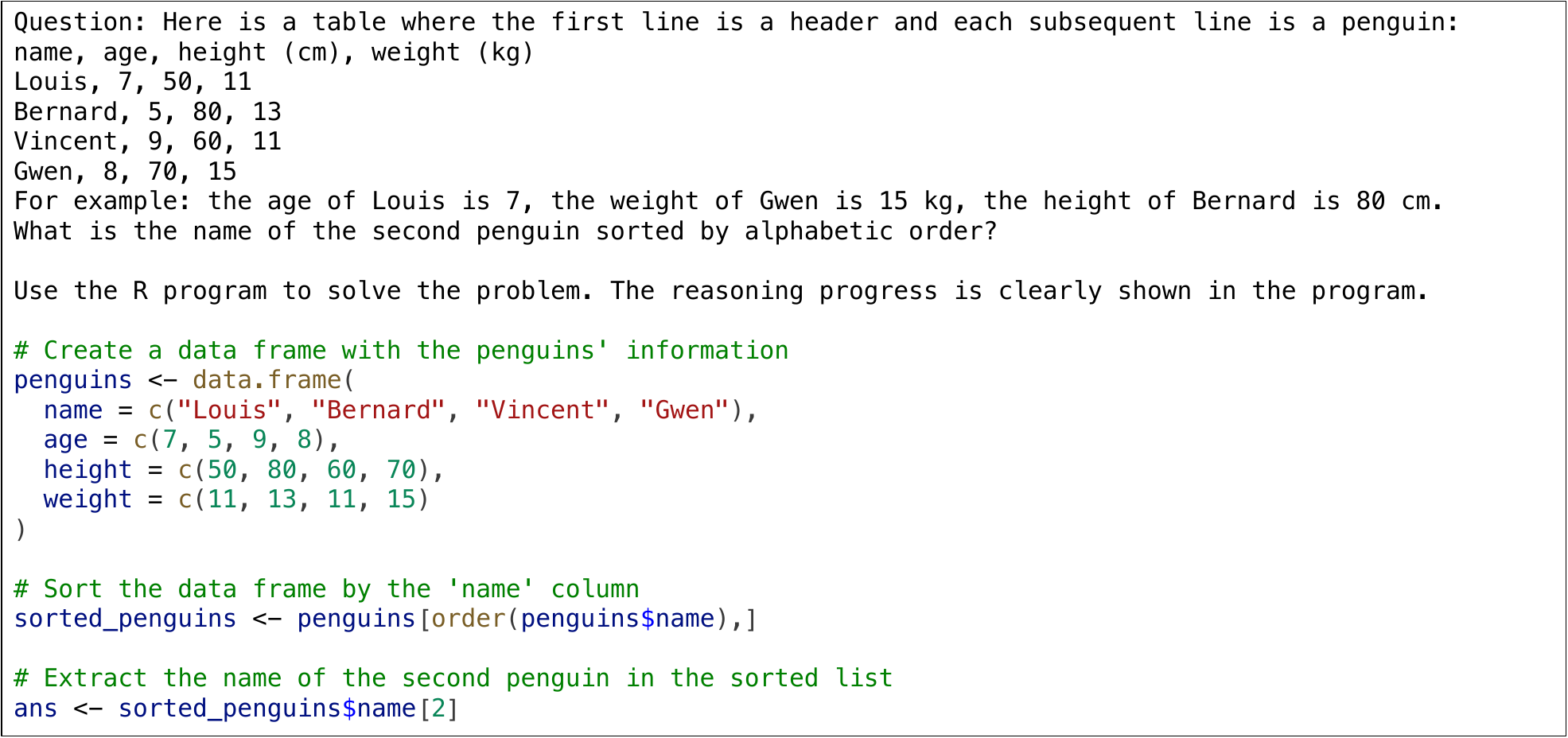}
\caption{R Prompt of the third question.}
\label{fig:cpp_prompt}
\end{figure*}

\end{document}